%% file: main.tex
\documentclass[10pt,twocolumn,letterpaper]{article}

\usepackage[pagenumbers]{cvpr} 

\input{preamble}
\definecolor{cvprblue}{rgb}{0.21,0.49,0.74}
\usepackage[pagebackref,breaklinks,colorlinks,allcolors=cvprblue]{hyperref}
\usepackage{multirow}
\usepackage[table]{xcolor}
\usepackage{colortbl}  
\usepackage[table,xcdraw]{xcolor}  
\usepackage{amsmath}

\def\paperID{} 
\def\confName{CVPR}
\def\confYear{2026}

\title{Video-CoE: Reinforcing Video Event Prediction via Chain of Events}

\author{
\hspace{-0.9cm}
\textbf{Qile Su}\textsuperscript{\textsection}\thanks{Work done during the internship at AMAP, Alibaba Group.},
\textbf{Jing Tang}\textsuperscript{$\ddagger$},
\textbf{Rui Chen}, 
\textbf{Lei Sun},
\textbf{Xiangxiang Chu}
\vspace{0.5em}
\\
{
AMAP, Alibaba Group\quad} \\
\vspace{0.5em}
{\small 
\textsuperscript{$\ddagger$} Project lead. \quad \textsuperscript{\textsection} Corresponding author.} \\
\vspace{-2em}
}

\begin{document} 
\maketitle
\input{sec/0_abstract}    
\input{sec/1_intro}
\input{sec/2_related}
\input{sec/2_5_evaluation}
\input{sec/3_method}
\input{sec/4_experiments}
\input{sec/5_conclusion}
{
    \small
    \bibliographystyle{ieeenat_fullname}
    \bibliography{main}
}

\input{sec/X_suppl}

\end{document}

%% file: sec/0_abstract.tex
\begin{abstract}
Despite advances in the application of MLLMs for various video tasks, video event prediction (VEP) remains relatively underexplored. VEP requires the model to perform fine-grained temporal modeling of videos and establish logical relationships between videos and future events, which current MLLMs still struggle with.
In this work, we first present a comprehensive evaluation of current leading MLLMs on the VEP task, revealing the reasons behind their inaccurate predictions, including lack of logical reasoning ability for future events prediction and insufficient utilization of visual information.
To address these challenges, we propose \textbf{C}hain \textbf{o}f \textbf{E}vents (\textbf{CoE}) paradigm, which constructs temporal event chains to implicitly enforce MLLM focusing on the visual content and the logical connections between videos and future events, incentivizing model's reasoning capability with multiple training protocols.
Experimental results on public benchmarks demonstrate that our method outperforms both leading open-source and commercial MLLMs, establishing a new state-of-the-art on the VEP task.
Codes and models will be released soon.

\end{abstract}

%% file: sec/1_intro.tex
\section{Introduction}
\label{sec:intro}


\begin{figure*}[t]    
  \centering            
  \subfloat[An example of MLLMs performing VEP.]
  {
      \label{fig:subfig1}\includegraphics[width=0.280\textwidth]{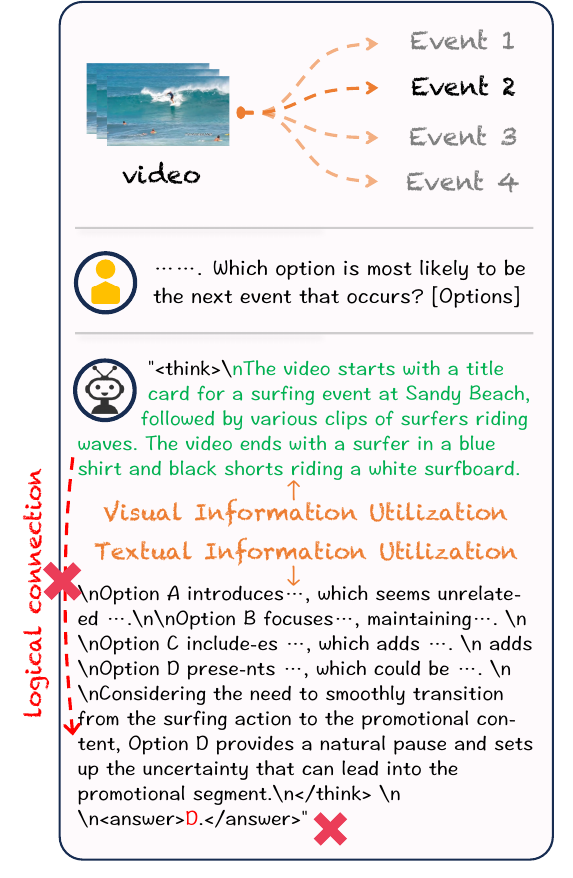}
  }
  \subfloat[Attention distribution over input token sequences.]
  {
      \label{fig:subfig2}\includegraphics[width=0.705\textwidth]{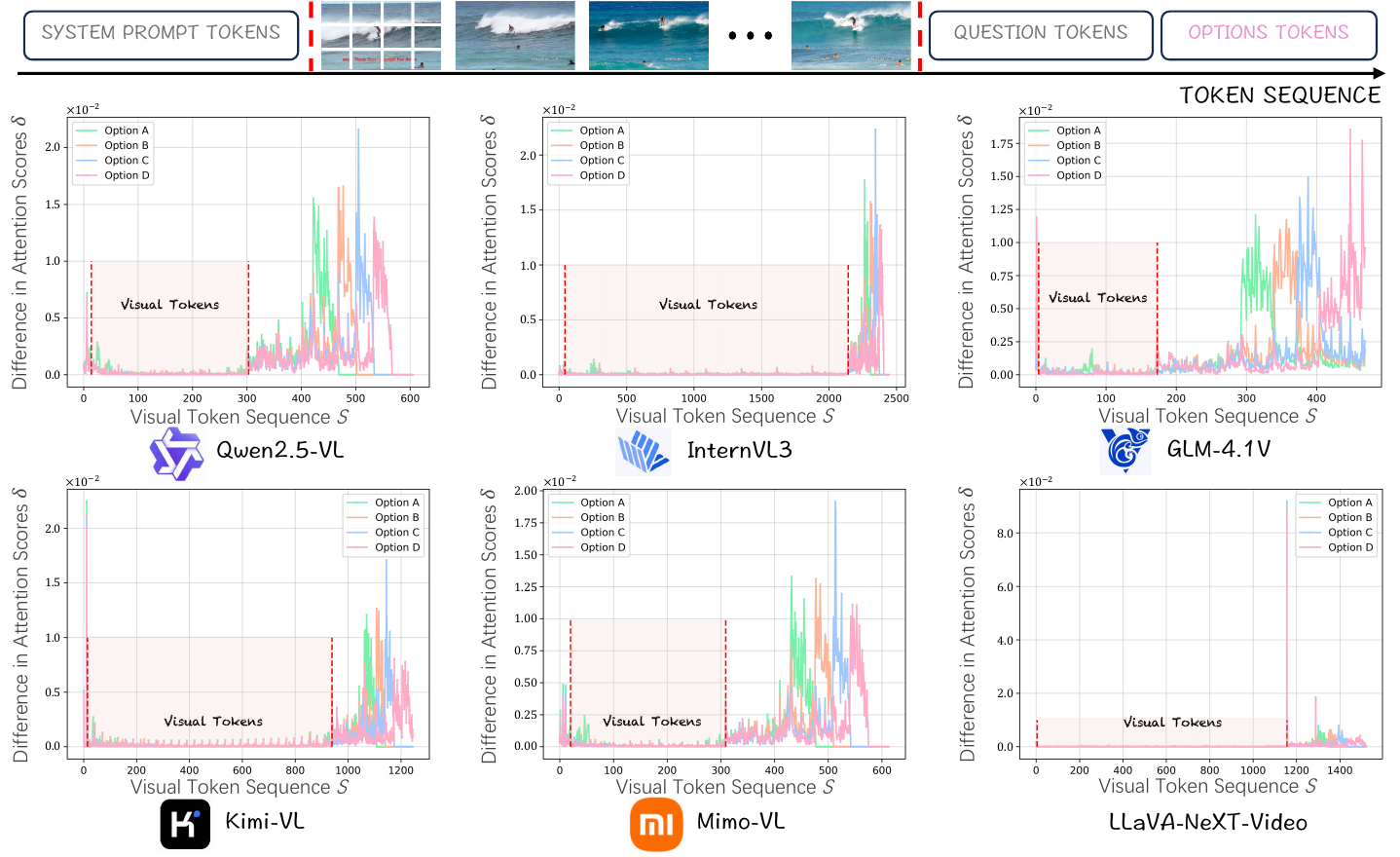}
  }
  \caption{Analysis of open-source MLLMs on video event prediction tasks. \cref{fig:subfig1} illustrates the reasoning process, indicating the lack of logical reasoning capabilities in VEP task. \cref{fig:subfig2} illustrates the attention distribution of the option tokens over input tokens demonstrating the insufficient utilization of visual information.}    
  \label{fig:mllms_analyse}            
\end{figure*}

Multimodal Large Language Models (MLLMs) \cite{gpt4o, internvl3_5, Qwen2.5-VL} have achieved remarkable results across a range of vision tasks \cite{VSI-Bench, mmvu, mvbench, tempcompass, video-mme, videommmu}, demonstrating strong capabilities in video understanding, reasoning, and question answering.
These tasks collectively underpin the predominant pre-training and post-training paradigms for MLLMs, enabling them to generalize effectively to diverse downstream applications~\cite{medical, education, safety}.
Nevertheless, real-world scenarios, such as crisis early warning, require the ability to predict future events from observed videos, a capability that remains largely underexplored in current MLLM research.

To fill this gap, we first conduct a systematic evaluation of state-of-the-art open-source MLLMs \cite{glm41v, internvl3_5, kimivl, mimovl, Qwen2.5-VL, zhang2024llavanextvideo} and commercial GPT-series models~\cite{gpt4o} on the video event prediction (VEP)~\cite{videvent,futurebench} task, as shown in \cref{tab:futurebench,tab:AVEP}.
Our experiments indicate that current MLLMs perform markedly worse on VEP than on standard vision tasks.
We attribute this gap to insufficient pretraining on the VEP task, which leaves models without the inductive biases and reasoning skills required for accurate future events prediction.
Directly training these models for VEP task would require large-scale datasets and substantial computational resources, making it costly to incorporate this objective into pretraining.
This motivates a more efficient approach to strengthen MLLMs’ video event prediction capabilities without large-scale annotation or extensive retraining.
%
%

To this end, we perform a systematic analysis of the limitations faced by state-of-the-art open-source MLLMs on zero-shot video event prediction.
As illustrated in \cref{fig:mllms_analyse}, our study uncovers two primary failure causes:
%

\textbf{Lack of Logical Reasoning Ability for Future Events.}
Unlike standard video understanding and reasoning tasks, VEP aims to anticipate plausible future events that are not directly observable in the input video.
This requires models to possess the ability to reason over the video content to predict the future events.
However, as shown in~\cref{fig:subfig1}, current MLLMs often rely on cues in textual answer options rather than grounding predictions in video evidence, indicating a weak linkage between observed content and the future.
This shortcut behavior contributes to their subpar performance on VEP.
Moreover, in real-world applications, video event prediction is inherently an open-set problem, where future events are not confined to a fixed label space, further limiting the practical applicability of current MLLMs.
%
%

\textbf{Insufficient Utilization of Visual Information.}
As shown in \cref{fig:subfig1}, our observations indicate that current MLLMs make limited use of visual evidence during reasoning, instead over-relying on textual cues or answer choices. 
%
An analysis of attention distributions over visual and textual tokens, shown in \cref{fig:subfig2}, further reveals that models allocate substantially less attention to visual tokens during prediction.
%
Yet prior studies \cite{granroth2016happens, li2018constructing, videvent} demonstrate that fine-grained temporal modeling is essential for forecasting future events. 
%
This text-centric modality bias likely undermines robust predictive reasoning, leading to suboptimal performance on VEP. 
%
Although previous works \cite{paying, see, prompt, ibd} have proposed (i) directly amplifying attention to visual tokens at inference and (ii) using prompts to encourage visual grounding, we find these approaches ineffective for VEP and even lead to performance degradation.
%
%

To address these challenges, we propose \textbf{Chain of Events (CoE)}, a paradigm for video event prediction.
%
CoE first constructs a fine-grained temporal representation by segmenting the input video into a sequence of historical events, forming an explicit event chain. 
%
This step promotes stronger visual grounding and mitigates the common visual–textual utilization bias in MLLMs, providing a more reliable basis for subsequent logical reasoning. 
%
The model then reasons jointly over the observed video and the constructed event chain to anticipate plausible future events, rather than relying on superficial cues from textual options analysis. 
%
By explicitly linking observed events to potential future events via causal–temporal reasoning, CoE enhances predictive performance on VEP task and directly addresses the limitations of current MLLMs.
%
%
%

To enforce the model adhering to our proposed CoE paradigm, we introduce a two-stage training approach, CoE-SFT and CoE-GRPO, which facilitates model's adaptation to the CoE framework and enhances video event prediction accuracy with modest training costs.
In stage one training, \textbf{CoE-SFT} fine-tunes the model through supervised learning, enforcing the model to establish logical connections between historical video evidence and future events during the reasoning process, rather than serving merely as a cold-start.
In the second stage, \textbf{CoE-GRPO} strengthens model's temporal localization and video understanding capabilities, enabling the model to construct fine-grained temporal event chains, providing sufficient visual information and logical support for prediction.
%
%

We evaluate our approach on established video event prediction benchmarks using Qwen2.5-VL~\cite{Qwen2.5-VL} as our base model and compare it with strong open-source and commercial MLLMs. 
Experimental results demonstrate that our method significantly enhances the utilization of visual information and enables logical reasoning over video content to predict future events, achieving state-of-the-art performance across various VEP benchmarks.
Furthermore, we validate the superiority of our approach in open-set prediction scenarios through an evaluation with a judge model. Our main contributions are as follows:
%
\begin{itemize}
    \item We propose an effective video event prediction paradigm, Chain of Events, which addresses the challenges faced by existing MLLMs in video event prediction and significantly improves their accuracy in predicting future events.
    \item We propose an efficient method to implement the CoE paradigm, which unlocks the MLLMs' ability to construct temporal event chains and enables them to reason over the observed video to predict future events logically.
    \item We establish one of the most comprehensive baselines to date for the VEP task through a systematic evaluation of our method and a wide range of MLLMs on this task, providing a solid foundation for future research in this area. Our experiments demonstrate that the proposed method effectively addresses the challenges faced by MLLMs in VEP, achieving SOTA performance across benchmarks.
\end{itemize}

%% file: sec/2_related.tex
\section{Related Works}

\subsection{Video Event Prediction}
The video event prediction (VEP) task was first introduced in~\cite{videvent}, which requires the model to predict the next possible event based on the input video.
Unlike other video reasoning tasks~\cite{tempcompass, temporalbench, seed-bench-r1, chen2025finger, clevr} focusing on the video content itself, VEP demands the model to infer unseen future content from current visible evidence, thereby posing higher requirements on model's video understanding and logical reasoning capabilities.
Previous works~\cite{granroth2016happens, li2018constructing, videvent, Zhu_Gao_Yu_Wang_Xu_Mu_Yang_Xu_2023} have shown that fine-grained temporal modeling of historical events is critical for accurately forecasting future events.
Thus, the concept of Event Chains~\cite{granroth2016happens} has been widely adopted as an effective temporal representation paradigm in event modeling for both textual~\cite{10.1145/3758126.3758132, 10.5555/3304222.3304354, 10.1145/3626772.3657706} and video event prediction tasks~\cite{videvent}.
Recent works (VidEvent~\cite{videvent}, AVEP~\cite{avep}, and NEP~\cite{futurebench}) have analyzed the performance of MLLMs on VEP tasks, indicating that existing methods failed to achieve satisfactory results.
However, 
no prior works have systematically investigated why MLLMs performed poorly in VEP tasks, nor have there been comprehensive evaluations or targeted methods to enhance their reasoning for future event prediction, particularly methods that enable large models to effectively model the evolution of historical events in videos.

\subsection{Visual Large Language Models for Reasoning}
With the rapid advancement of MLLMs’ video understanding capabilities \cite{gpt4o, Qwen2.5-VL, internvl3_5, videoxl, apollo, wang2026urban, huang2025taming, yuan2026if} and LLMs’ reasoning abilities \cite{deepseek, qwen3, glm2024chatglm, token, logicrl}, recent studies have increasingly focused on exploring the reasoning capabilities of MLLMs \cite{peng2025skyworkr1vpioneeringmultimodal, videor1, visualrft, grit, videochata1,wang2026urban}.
Several models, such as Qwen2.5-VL \cite{Qwen2.5-VL}, GLM-4.1V \cite{glm41v}, Kimi-VL \cite{kimivl}, have been trained on various visual reasoning tasks, achieving competitive results and demonstrating great potential.
Beyond supervised training, many works \cite{kimivl, visionr1, logicrl, chen2025reasoningerasurveylong, grit} have followed the GRPO approach proposed by DeepSeek-R1 \cite{deepseek}, leveraging RL to further enhance reasoning capabilities.
For instance, Open-Reasoner \cite{openreasoner}, Kimi-VL \cite{kimivl}, and Mimo \cite{mimovl} adopt similar RL pipelines to strengthen reasoning performance.
Building upon GRPO, several works \cite{videochat, lover1, videorft, twgrpo, chu2026gpg, videor1, grit,li2025adacurl,yuan2026autodriver, yuan2026videostar, bai2025univg} have proposed adaptive modifications to further enhance the performance of MLLMs on visual reasoning tasks. 
However, these methods primarily focus on frame-level or local-region perception and are not tailored for event prediction.
In the context of VEP, NEP \cite{futurebench} demonstrates that standard GRPO \cite{deepseek} training yields better performance than standard SFT.
However, despite these promising advances, MLLMs’ performance on VEP remains largely underexplored. And there is still a lack of targeted methods specifically designed to enhance their event prediction capabilities.

%% file: sec/2_5_evaluation.tex
\section{Evaluation and Analysis of MLLMs on VEP}
We conduct a systematic evaluation of various open-source and commercial MLLMs on the VEP task, as shown in \cref{tab:futurebench,tab:AVEP}. The results indicate that current MLLMs do not exhibit the same strong performance on VEP as they do in other vision tasks. Among them, \textbf{Qwen3-VL} demonstrates the best performance across most metrics, yet the average accuracy is only $66.9\%$. The results suggest that MLLMs still have significant room for improvement in VEP, highlighting the research potential. 

As shown in \cref{fig:subfig1}, we visualized the model's reasoning process and found that existing MLLMs generally follow a pattern: they first generate a high-level description of the video, then analyze each option, and finally select the most relevant option as the correct answer. This reasoning process lacks the logical connection between the video and future events, meaning the model does not truly reason about future events from the video but rather chooses the most relevant option. This often leads to incorrect predictions. 

Additionally, as shown in \cref{fig:subfig1}, we observed that the model tends to generate coarse-grained summaries of the video, which may cause it to overlook critical details relevant to future events and neglect the temporal dynamics underlying event evolution. Throughout the reasoning process, the utilization of visual information is significantly lower than that of textual information. 
We further investigate the attention distribution of MLLMs when performing the VEP task. Due to the causal attention mechanism, later tokens inherently contain information from earlier tokens. To avoid interference from this effect, we visualized the attention distribution specifically over the input option tokens, which also provides a fair comparison of attention patterns across different models by mitigating the influence caused by differences in generated tokens.
As shown in \cref{fig:subfig2}, we found that the attention distribution on visual tokens is much lower than that on textual tokens, indicating that the model fails to adequately focus on and utilize visual information when performing the VEP task.

Based on these experiments, we conclude that there is considerable room for improvement in the performance of MLLMs on VEP. The key challenges lie in addressing the lack of logical reasoning ability for future events and the insufficient utilization of visual information.

%% file: sec/3_method.tex
\section{Method}
In the following section, we will provide a detailed overview of the CoE paradigm and how the CoE-SFT and CoE-GRPO are employed to implement it.

\subsection{Chain of Events (CoE) Paradigm}
Previous works \cite{nguyen2025hyperglmhypergraphvideoscene, wang2025videotreeadaptivetreebasedvideo} often use structured representations such as chains, trees, and graphs for video modeling. However, they are mostly action-centric and designed for localization or understanding tasks, while overly complex representations introduce unnecessary learning overhead for MLLMs. Therefore, we propose a CoE paradigm to model historical events in a fine-grained manner.

We define the model's reasoning process as $\mathcal{R}=\text{MLLM}^{reason}(V,Q)$, in which $V$ denotes the input video, $Q$ denotes the question. And the VEP process of the vanilla model is then expressed as:
\begin{equation}
    P=P(\hat{E}\mid V,Q,\mathcal{R}),
\end{equation}
\begin{figure}[t]
    \centering
    \includegraphics[width=1.0\linewidth]{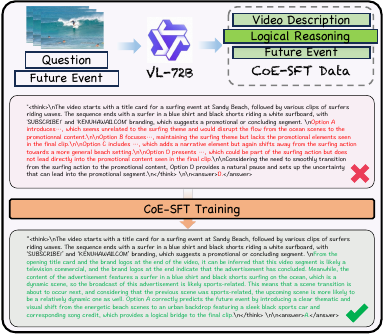}
    \caption{An illustration of our proposed \textbf{CoE-SFT} method within Qwen2.5-VL-72B. We provide the larger model with the video and the future event, and prompt it to generate the intermediate logical reasoning process that connects them. Training on such data encourages the model to develop logical reasoning abilities rather than relying on option-based analysis.}
    \label{fig:coe-sft}
\end{figure}
where $\hat{E}$ denotes the predicted event.
In the CoE paradigm, we define an event $E$ as a pair $E=(\mathcal{T},\mathcal{D})$, where $\mathcal{T}$ denotes the start and end timestamps of a video event and $\mathcal{D}$ denotes the textual description of the event.
A temporal event chain $EC$, therefore, can be defined as a sequence of events occurring in the video, ordered temporally, $EC=[E_1,E_2,\dots,E_n]$. 
Consequently, this paradigm can be formalized as follows: The model first performs fine-grained temporal modeling of the video to construct the event chain $EC=\text{MLLM}^{CoE}(V)$. Then, it reasons based on the video content and the event chain $\mathcal{R'}=\text{MLLM}^{reason}(V,Q,EC)$, where the $\mathcal{R'}$ incorporates logical reasoning from video content to future events, given the specific nature of the event prediction task. Finally, the model’s prediction process can be expressed as:
\begin{equation}
   P=P(\hat{E}\mid V,Q,\mathcal{R'},EC).
    \label{eq:1}
\end{equation}

The CoE paradigm addresses the limitations faced by MLLMs in VEP through two mechanisms: (i) a reasoning process that incorporates the logical connections between video content and the future event, and (ii) fine-grained temporal modeling via the construction of event chains.
\begin{figure*}[t]
    \centering
    \includegraphics[width=1.0\linewidth]{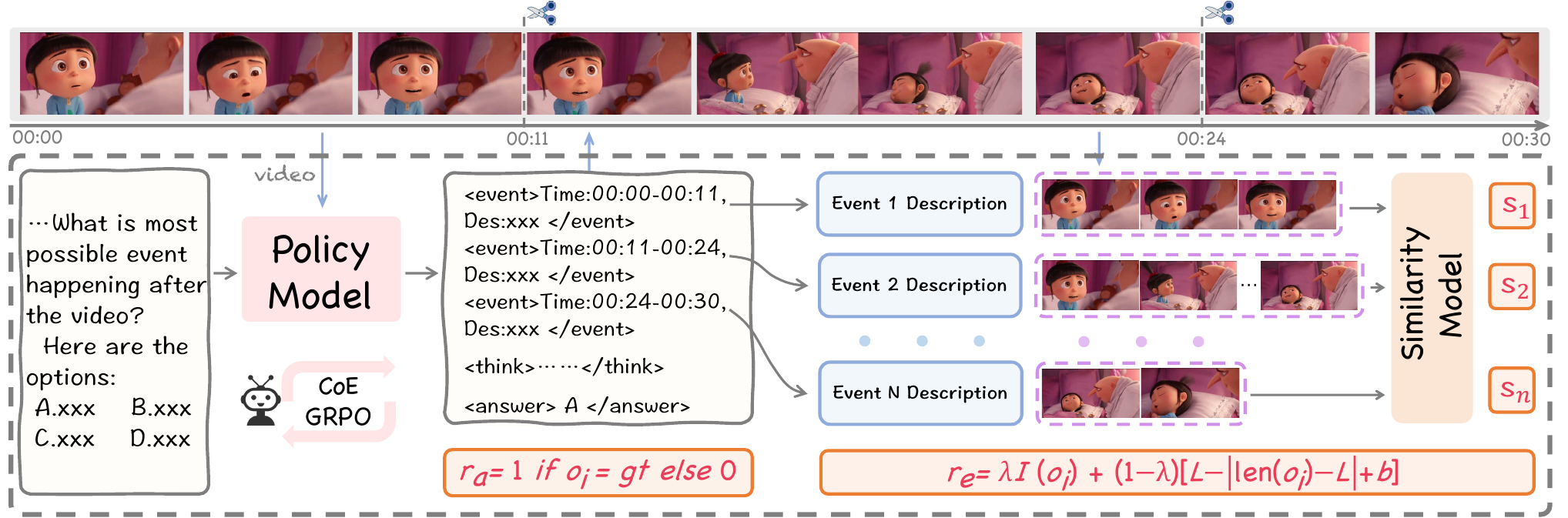}
    \caption{An illustration of our proposed \textbf{CoE-GRPO} method. The overall supervision signal consists of three components: $r_e$ encourages the model to follow the CoE reasoning paradigm and constrains the CoE length; $r_s$ supervises the alignment between event timestamps and textual descriptions while preventing reward hacking; and $r_a$ provides verifiable reward signals. The scissor icon indicates the temporal segmentation of video clips based on timestamps.}
    \label{fig:method}
\end{figure*}
\subsection{CoE with Supervised Fine-Tuning}
Unlike other video tasks, VEP requires models to predict unobservable future events based on the video content. This necessitates the model constructing a logical reasoning process between the observed video content and the unobserved future events. However, existing \textbf{vanilla SFT} data \cite{futurebench} is typically constructed by sequentially analyzing answer options, which fails to address the absence of a logical reasoning process. Consequently, despite fine-tuning on datasets of over 30K samples, the performance improvements remain limited \cite{futurebench}. To address this, following the CoE paradigm, we propose the CoE-SFT method, which focuses on constructing the logical connections between the video and future events during the reasoning process. 

Specifically, as shown in \cref{fig:coe-sft}, we utilize a powerful large model, Qwen2.5-VL-72B, to construct a small-scale CoE-SFT training dataset. We provide the video, question, and correct future event to the model, and instruct it to output the logical reasoning process that derives the future event from the video content, while avoiding any analysis of other options. Following this, we perform a manual quality check to ensure data quality, achieving a pass rate of over $90\%$.
It is worth noting that we did not construct the $EC$ in the CoE-SFT data, as the quality of $EC$ construction from large-scale MLLMs did not meet expectations and could potentially hinder the model's training. However, we observed that the model still effectively retained its reasoning ability based on video content and achieved the expected results after CoE-GRPO training. We have provided examples of the model's reasoning process in the Supplementary Material.

To better assess the model's logical reasoning ability and its performance in real-world applications, we propose an open-set \textbf{judge model evaluation} metric. In this evaluation, the judge model assesses both the validity of reasoning and the correctness of the answers, selecting the best response from multiple competing models and providing the reason behind its choice. The final win rate is then used as the evaluation metric.

\subsection{CoE with Group Relative Policy Optimization}
The foundation of event prediction lies in temporal modeling of historical events \cite{granroth2016happens, li2018constructing, videvent, Zhu_Gao_Yu_Wang_Xu_Mu_Yang_Xu_2023}. However, the insufficient utilization of visual information in MLLMs during event prediction hinders their ability to perform fine-grained temporal modeling, resulting in suboptimal performance. 

To address this limitation, we propose the CoE-GRPO, an improved GRPO framework specifically designed for VEP. Our method effectively unlocks the model's temporal localization and video understanding capabilities for constructing event chains, enabling fine-grained temporal modeling and improving the model's utilization of visual information during event prediction. 

Specifically, as illustrated in \cref{fig:method}, we first introduce the special event tags \texttt{<event>} and \texttt{</event>} to explicitly mark the boundaries of event $E$ within the model output. Each event tag pair contains the start and end timestamps of a corresponding event, $t_{start}$ and $t_{end}$, as well as a fine-grained description $\mathcal{D}$ capturing its semantic details:
\begin{equation}
\small
    E=\texttt{<event>Time:}t_{start}-t_{end}\texttt{,Des:}\mathcal{D} \texttt{</event>}.
\end{equation}
During the CoT reasoning process, the model incrementally constructs a historical event chain $EC$ consisting of multiple event tags organized in chronological order, which provides the visual grounding for subsequent logical reasoning steps. 
Since this relatively simple event representation method does not require additional data for cold start, we can directly employ reinforcement learning to train the model to construct event chains and leverage them for event prediction, as shown in \cref{eq:1}.

To achieve this, we introduce a targeted, dense \textbf{CoE reward} $r_{e}$, which provides fine-grained supervision throughout the model’s event chain construction process, allowing control over both the correct construction and the length of the event chain:
\begin{equation}
    r_{e}^{(i)}=\lambda I(o_i)+(1-\lambda)[L-\lvert \text{len}(o_i)-L\rvert +b].
\end{equation}
where $\lambda$ denotes the weight coefficient.
In this, the indicator function $I(o_i)$ takes the value 1 if $o_i$ correctly contains all the required tag tokens 
, and 0 otherwise. Since both excessively long and short event chains hinder the model’s ability, according to our experiments, we introduce a length constraint term to control the length of the output event chain. The function $\text{len}(o_i)$ calculates the number of events in the event chain $EC$ of completion $o_i$. Here, $L$ is a hyperparameter representing the model's ideal output length, and $b$ is a bias term used to ensure that the maximum value of $r_{e}$ is 1.

To ensure the consistency between the event descriptions and the video content in $E$, and to prevent the model from cheating by optimizing for rewards, we introduce a continuous \textbf{similarity reward} $r_s$ for supervision. Specifically, as shown in the \cref{fig:method}, we crop the original video according to the start and end timestamps of each event $E$ in the event chain $EC$ output by the model, obtaining a set of video clips $[clip_1,clip_2,\dots,clip_n]$, which corresponds to the event chain. We then compute the cross-modal similarity $s_j$ between the event description and the video clips. The average of these similarity values is used as the similarity reward signal, ensuring that the model constructs event chains that align closely with the videos:
\begin{equation}
    r_s=\frac{1}{n}\sum_{j=1}^n{s_j},
\end{equation}
in which $s_j=\text{cos}(v_j,t_j)$, $v_j$ and $t_j$ are the visual and textual features of the event embedded by the similarity model. We use different similarity models and present their performance in the Experiment section, with the details of similarity calculation provided in the Supplementary Material.

Based on the aforementioned reward signals, the final reward is computed as the weighted sum of the individual reward components:
\begin{equation}
    r_i=\alpha r^{(i)}_{a}+\beta r^{(i)}_{e}+(1-\alpha-\beta)r^{(i)}_{s},
\end{equation}
where $r_a$ serves as the accuracy reward, $\alpha$ and $\beta$ denote the reward weights. During training, we sample a group of $N$ completions from the current policy $\pi_\theta$. For each completion, we compute a reward $r_i$. The advantage $A_i$ is then calculated by normalizing the rewards within the group:
\begin{equation}
    A_i=\frac{r_i-\text{mean}(\{r_i\})}{\text{std}(\{r_i\})+\delta},
\end{equation}
where $\delta$ is a small constant for numerical stability.
Following DeepSeek-R1 \cite{deepseek}, the final policy update is as follows:
\begin{equation}
    \begin{split}
        \mathcal{J}(\theta )&=E_{q,\{o_i\}} [\frac{1}{G}\sum_{i=1}^{G}(\text{min}(r_{ratio}A_i,
        \text{clip}(r_{ratio},\\ &1-\epsilon,1+\epsilon)A_i)-\beta D_{KL}(\pi_{\theta}||\pi_{ref})],
    \end{split}
\end{equation}
where the importance sampling ratio $r_{ratio}=\frac{\pi_{\theta}(o_i\mid q)}{\pi_{\theta_{old}}(o_i\mid q)}$.

CoE-GRPO can efficiently unlock the model's temporal localization and video understanding capabilities, enabling fine-grained temporal modeling of historical videos through event chain construction. This enhances visual information utilization and improves event prediction accuracy. Additionally, the method leverages the model's inherent capabilities without the need for additional data annotations, making it en efficient approach.


\subsection{Training}
We use Qwen2.5-VL-3B/7B as the base MLLMs. We adopt the two-phase training approach: CoE-SFT followed by CoE-GRPO.
In the first phase, CoE-SFT, we train the model using the small-scale CoE-SFT reasoning dataset, enabling the model to logically reason from visual content to infer the potential future events. The resulting model is denoted as \textbf{CoE-SFT}.
In the second phase, CoE-GRPO, we continue to train the model using the proposed method on the RL datasets across various benchmarks. This phase focuses on training the model to achieve fine-grained temporal modeling by constructing event chains. The resulting model is denoted as \textbf{CoE-GRPO}.

%% file: sec/4_experiments.tex
\section{Experiments}
\begin{table*}[]
\centering
\caption{Evaluation results of open-source/commercial MLLMs and our proposed \textbf{CoE} on FutureBench~\cite{futurebench}.}
\label{tab:futurebench}
\renewcommand{\arraystretch}{0.6}
\setlength{\tabcolsep}{11.5pt}
\small
\begin{tabular}{l|c|c|ccccc}
\toprule
\multirow{2}{*}{Model} & \multirow{2}{*}{Method} & \multirow{2}{*}{Frames} & \multicolumn{5}{c}{Futurebench $\uparrow$} \\ \cline{4-8}
\rule{0pt}{2.0ex}& & & 1-Hop & 2-Hop & 3-Hop & Interp. & AVG \\ 
\midrule
GLM-4.1V-9B \cite{glm41v} & \multirow{11}{*}{Vanilla} & 32 & 29.9 & 41.9 & 52.2 & 47.3 & 44.38 \\
LLavA-NeXT-Video \cite{zhang2024llavanextvideo} &  & 32 & 48.8 & 49.3 & 40.0 & 44.4 & 45.24 \\
Kimi-VL-A3B \cite{kimivl} &  & 32 & 44.3 & 42.8 & 51.3 & 51.9 & 48.87 \\
MiMo-VL-7B \cite{mimovl} &  & 32 & 59.0 & 59.6 & 50.5 & 43.8 & 50.45 \\
InternVL3-8B \cite{internvl3_5} &  & 32 & 54.3 & 58.0 & 63.2 & 54.4 & 56.72 \\
Qwen2.5-VL-32B \cite{Qwen2.5-VL} &  & 32 & 66.5 & 62.7 & 63.2 & 55.2 & 59.94 \\ 
Qwen2.5-VL-72B \cite{Qwen2.5-VL} &  & 32 & 55.5 & 68.4 & 63.7 & 53.2 & 58.33 \\
Qwen3-VL-30B-A3B \cite{qwen3} &  & 32 & 65.3 & 70.5 & \textbf{76.1} & 62.2 & 66.86 \\
GPT-4o \cite{gpt4o} &  & 32 & 61.9 & 61.7 & \underline{72.1} & 51.6 & 59.04 \\
GPT-5 & & 32 & 59.6 & 57.3 & 62.6 & 55.6 & 57.92 \\
\midrule
\midrule
\multirow{5}{*}{Qwen2.5-VL-3B} & Instruct \cite{Qwen2.5-VL} & 32 & 45.1 & 50.8 & 44.3 & 45.8 & 46.30 \\
& (NEP) SFT \cite{futurebench} & 32 & 53.2 & 57.0 & 59.2 & 59.3 & 57.86 \\
& \cellcolor[rgb]{0.85, 1.0, 1.0}CoE-SFT (Ours) & \cellcolor[rgb]{0.85, 1.0, 1.0}32 & \cellcolor[rgb]{0.85, 1.0, 1.0}68.8 & \cellcolor[rgb]{0.85, 1.0, 1.0}\underline{75.1} & \cellcolor[rgb]{0.85, 1.0, 1.0}55.7 & \cellcolor[rgb]{0.85, 1.0, 1.0}60.5 & \cellcolor[rgb]{0.85, 1.0, 1.0}63.60 \\
& (NEP) GRPO \cite{futurebench} & 32 & 63.0 & 65.3 & 63.7 & 67.1 & 65.45 \\
& \cellcolor[rgb]{1.0, 0.85, 0.85}CoE-GRPO (Ours) & \cellcolor[rgb]{1.0, 0.85, 0.85}32 & \cellcolor[rgb]{1.0, 0.85, 0.85}\underline{71.1} & \cellcolor[rgb]{1.0, 0.855, 0.85}73.6 & \cellcolor[rgb]{1.0, 0.85, 0.85}69.7 & \cellcolor[rgb]{1.0, 0.85, 0.85}64.6 & \cellcolor[rgb]{1.0, 0.85, 0.85}\underline{68.28}\\ 
\midrule
\multirow{5}{*}{Qwen2.5-VL-7B} 
& Instruct \cite{Qwen2.5-VL} & 32 & 57.2 & 57.0 & 50.2 & 50.7 & 52.94 \\ 
& (NEP) SFT \cite{futurebench} & 32 & 60.7 & 74.2 & 66.2 & 65.0 & 64.39 \\ 
& \cellcolor[rgb]{0.85, 1.0, 1.0}CoE-SFT (Ours) & \cellcolor[rgb]{0.85, 1.0, 1.0}32 & \cellcolor[rgb]{0.85, 1.0, 1.0}67.6 & \cellcolor[rgb]{0.85, 1.0, 1.0}74.1 & \cellcolor[rgb]{0.85, 1.0, 1.0}62.2 & \cellcolor[rgb]{0.85, 1.0, 1.0}63.2 & \cellcolor[rgb]{0.85, 1.0, 1.0}65.72 \\ 
& (NEP) GRPO \cite{futurebench} & 32 & 66.2 & 69.9 & 63.7 & \underline{68.1} & 67.28 \\ 
& \cellcolor[rgb]{1.0, 0.85, 0.85}CoE-GRPO (Ours) & \cellcolor[rgb]{1.0, 0.85, 0.85}32 & \cellcolor[rgb]{1.0, 0.85, 0.85}\textbf{80.9} & \cellcolor[rgb]{1.0, 0.85, 0.85}\textbf{83.9} & \cellcolor[rgb]{1.0, 0.85, 0.85}71.6 & \cellcolor[rgb]{1.0, 0.85, 0.85}\textbf{71.4} & \cellcolor[rgb]{1.0, 0.85, 0.85}\textbf{75.00} \\
\bottomrule

\end{tabular}
\end{table*}

\begin{table*}[]
\centering
\caption{Evaluation results of open-source MLLMs and our proposed \textbf{CoE} on AVEP~\cite{avep}.}
\label{tab:AVEP}
\renewcommand{\arraystretch}{1.0}
\setlength{\tabcolsep}{4pt}
\small
\begin{tabular}{l|>{\centering\arraybackslash}p{2.5cm}cccccc}
\toprule
\multirow{2}{*}{Method} & \multirow{2}{*}{Verb (Test / Val) $\uparrow$} & \multicolumn{3}{c}{Noun (Test / Val) $\uparrow$} & \multicolumn{3}{c}{Action (Test / Val) $\uparrow$} \\ \cline{3-8}
\rule{0pt}{2.0ex}&& Precision & Recall & F1-Score & Precision & Recall & F1-Score \\ \midrule 
LLaVA-Video-7B \cite{llavavideo} & 5.67 / 6.27 & 44.50 / 44.96 & 41.95 / 41.64 & 43.19 / 43.24 & 3.13 / 3.57 & 2.95 / 3.30 & 3.04 / 3.43 \\
Kimi-VL-A3B & 5.12 / 5.30 & 31.79 / 31.46 & 36.43 / 37.04 & 33.95 / 34.02 & 2.40 / 2.83 & 3.27 / 3.16 & 2.77 / 2.99 \\
InternVL3-8B & 5.80 / 7.00 & 48.77 / 43.64 & 45.77 / 37.94 & 47.22 / 40.59 & 3.37 / 3.60 & 3.23 / 3.07 & 3.30 / 3.31 \\
GLM-4.1V-9B & 9.10 / 10.62 & 33.80 / 34.94 & 33.66 / 35.42 & 33.73 / 35.18 & 3.01 / 3.27 & 3.02 / 3.34 & 3.01 / 3.30 \\
MiMo-VL-7B & \underline{10.23} / 9.57 & 43.46 / 39.54 & 32.07 / 35.10 & 36.91 / 37.19 &	\textbf{6.96} / 5.54 & 5.08 / 3.86 & 5.87 / 4.55 \\
Qwen2.5-VL-7B & 5.73 / 6.77 & 44.83 / 44.26 & 56.54 / 59.33 & 50.01 / 50.70 & 2.21 / 3.76 & 3.21 / 4.64 & 2.62 / 4.15 \\
Qwen2.5-VL-72B & 8.00 / 7.95 & \underline{48.94} / \textbf{49.24} & 41.38 /41.57 & 44.84 / 45.08 & 5.32 / 5.38 & 4.25 / 4.41 & 4.72 / 4.85 \\
Qwen2.5-VL-7B-GRPO \cite{deepseek} & 9.64 / 10.42 & 47.79 / 47.35 & \underline{90.07} / \underline{91.35} & \underline{62.45} / \underline{63.49} & 5.00 / \underline{5.80} & \underline{9.21} / \underline{9.72} & \underline{6.48} / \underline{7.26} \\
\cellcolor[rgb]{0.85, 1.0, 1.0}CoE-SFT-7B (Ours) & \cellcolor[rgb]{0.85, 1.0, 1.0}9.84 / \underline{11.44} & \cellcolor[rgb]{0.85, 1.0, 1.0}45.57 / 43.18 & \cellcolor[rgb]{0.85, 1.0, 1.0}54.74 / 59.02 & \cellcolor[rgb]{0.85, 1.0, 1.0}49.74 / 49.87 & \cellcolor[rgb]{0.85, 1.0, 1.0}4.94 / 5.72 & \cellcolor[rgb]{0.85, 1.0, 1.0}5.14 / 6.27 & \cellcolor[rgb]{0.85, 1.0, 1.0}5.04 / 5.98 \\
\cellcolor[rgb]{1.0, 0.85, 0.85}CoE-GRPO-7B (Ours) & \cellcolor[rgb]{1.0, 0.85, 0.85}\textbf{12.24} / \textbf{18.75} & \cellcolor[rgb]{1.0, 0.85, 0.85}\textbf{49.88} / \underline{48.68} & \cellcolor[rgb]{1.0, 0.85, 0.85}\textbf{93.95} / \textbf{93.54} & \cellcolor[rgb]{1.0, 0.85, 0.85}\textbf{65.16} / \textbf{64.03} & \cellcolor[rgb]{1.0, 0.85, 0.85}\underline{6.62} / \textbf{7.70} & \cellcolor[rgb]{1.0, 0.85, 0.85}\textbf{11.09} / \textbf{13.8} & \cellcolor[rgb]{1.0, 0.85, 0.85}\textbf{8.29} / \textbf{9.88} \\
\bottomrule
\end{tabular}
\end{table*}

\subsection{Setup}
\textbf{Benchmarks and Metrics.} We evaluate our method on public VEP benchmarks: FutureBench~\cite{futurebench} and AVEP~\cite{avep}. 
FutureBench measures the overall event prediction accuracy, where 1-HOP, 2-HOP, 3-HOP, and Interp. denote different prediction types, and AVG represents the overall average.
AVEP provides an evaluation of the prediction accuracy for event components, including verbs and event participants. Verb denotes the accuracy of event verb prediction, while the Precision, Recall, and F1-Score of Noun and Action respectively measure the precision, recall, and F1-Score of event participants and overall event predictions.
Since AVEP does not provide data for SFT, we do not include comparisons with vanilla SFT in our experiments on this benchmark. 
Details of these benchmarks and their evaluation metrics can be found in the Supplementary Material. \\
\textbf{Implementation Details.} We train our models using up to 16 NVIDIA H20 GPUs. For efficiency
considerations, we limit the maximum number of video frames to $32$ and the max resolution to $128\times28\times28$ during training. The GRPO group size $G$ is set to $4$, KL coefficient $\beta$ is set to $0.04$. The clipping parameter $\epsilon$ is set to $0.2$, and the learning rate is set to $1e-6$. We train our model for $150$ steps.

\begin{figure}[t]
    \centering
    \includegraphics[width=1.0\linewidth]{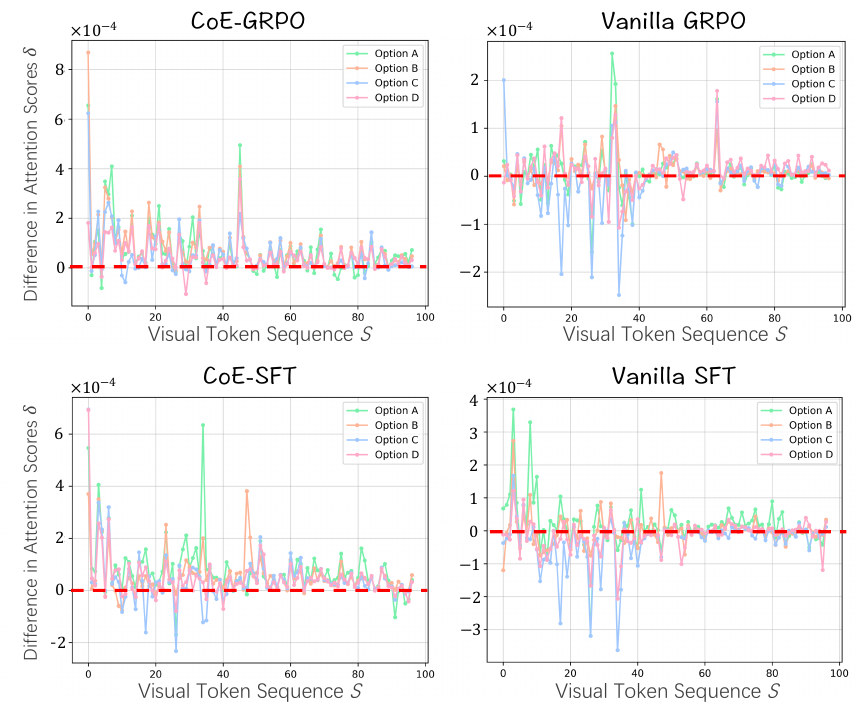}
    \caption{Attention difference of visual tokens between different methods and the base model. Portions greater than 0 indicate an improvement in attention.}
    \label{fig:attn_diff}
\end{figure}

\subsection{Main Results}
Results in \cref{tab:futurebench,tab:AVEP} indicate that \textbf{CoE-GRPO} consistently surpasses all baseline MLLMs on both benchmarks, delivering the best overall performance and highlighting the method’s effectiveness.
CoE-SFT achieves superior performance relative to vanilla SFT \cite{futurebench}, indicating that establishing logical connections between the video and future events helps improve prediction accuracy. CoE-GRPO also markedly outperforms vanilla GRPO \cite{deepseek,futurebench}, validating that modeling historical event chains improves the model’s capacity to predict future events.

Moreover, we empirically demonstrate that the proposed method addresses key VEP limitations of existing MLLMs. As shown in \cref{tab:attention,fig:attn_diff}, it substantially increases attention to visual tokens, whereas vanilla SFT \cite{futurebench} even reduces such attention. CoT~\cite{wei2022chain} refers to using prompts to guide model reasoning. Qualitative results in the Supplementary Materials further illustrate the improved utilization of visual information and the accurate logical connection between video and future events during reasoning.


In \cref{tab:judge}, judge-based evaluation results based on Qwen2.5-VL-72B show that CoE-SFT obtains the highest win rate. The small deficit for CoE-GRPO relative to CoE-SFT reflects the judge model’s familiarity with SFT-style reasoning rather than the CoE paradigm. Even so, their closely matched performance demonstrates that CoE-GRPO effectively preserves the logical reasoning needed for event prediction. To ensure fairness, we manually inspected the judge model’s evaluation process. The complete judge outputs are provided in the Supplementary Material.

\subsection{Ablation Study}
\textbf{Visual Attention Enhancement Methods.}
As summarized in \cref{tab:ablation}, we evaluate two common strategies for boosting visual attention. The first, Prompt-guided, instructs the model via a prompt to produce a detailed video description. The second, Constant-Bias, adds a fixed value to the attention weights of visual tokens at inference. Despite testing multiple prompt formulations and bias magnitudes, both strategies yielded performance degradation.
\\ \textbf{Ablation of Group Size.} The group size $G$ is a hyperparameter that controls the number of rollouts during CoE-GRPO. We evaluate the training results with different values of $G$, as shown in the \cref{tab:ablation}. The results indicate that the model's performance improves as $G$ increases. However, an excessive number of rollouts leads to higher training costs. Therefore, we recommend setting it to 4 to achieve a favorable balance between performance and training cost. 
\begin{table}[t]
\centering
\caption{Attention variation of visual tokens. WR (Winning Rate) denotes the overall percentage of samples showing attention improvement compared to the base model, while IR (Improvement Rate) indicates the average numerical increase in attention weight compared to the base model.}
\label{tab:attention}
\renewcommand{\arraystretch}{1.0}
\setlength{\tabcolsep}{5pt}
\small
\begin{tabular}{c|ccccc}
\toprule
Method & SFT & CoT & GRPO & CoE-GRPO & CoE-SFT \\
\hline
Baseline & \multicolumn{5}{c}{Qwen2.5-VL-7B-Instruct} \\
\midrule
WR $\uparrow$& 0.32 & 0.44 & 0.59 & \underline{0.77} & \textbf{0.93} \\
\midrule
IR(\%) $\uparrow$& -3.33 & +1.08 & +1.47 & \underline{+9.20} & \textbf{+15.11} \\
\bottomrule
\end{tabular}
\end{table}
\begin{table}[h]
\centering
\caption{The results of judge evaluation of different method. WR represents the ratio of samples where the model's answers are more reasonable and accurate compared to other models.}
\label{tab:judge}
\renewcommand{\arraystretch}{0.6}
\setlength{\tabcolsep}{4pt}
\small
\begin{tabular}{c|ccccc}
\toprule
Method & CoE-SFT& CoE-GRPO & Instruct & SFT & GRPO \\
\midrule
WR(\%) $\uparrow$& \textbf{38.13} & \underline{32.42} & 16.21 & 7.88 & 5.37 \\
\bottomrule
\end{tabular}
\end{table}
\begin{table}[t]
\centering
\caption{The results of the ablation studies.}
\label{tab:ablation}
\renewcommand{\arraystretch}{0.8}
\setlength{\tabcolsep}{4pt}
\small
\begin{tabular}{cccccc}
\toprule
Item & 1-Hop & 2-Hop & 3-Hop & Interp. & AVG \\
\midrule
\multicolumn{6}{c}{\textit{Visual Attention Enhancement Methods}} \\
\midrule
Prompt-guided & 44.5 & 46.6 & 43.8 & 46.6 & 45.74 \\
Constant-Bias & 54.6 & 52.3 & 57.6 & 50.6 & 52.57 \\
CoE (Ours) & \textbf{80.9} & \textbf{83.9} & \textbf{71.6} & \textbf{71.4} & \textbf{75.00} \\
\midrule
\multicolumn{6}{c}{\textit{Group Size} $G$} \\
\midrule
$G=2$ & 57.8 & 64.8 & 60.7 & 59.9 & 60.61 \\
$G=4$ & 77.5 & 78.2 & 71.6 & 73.4 & 74.61 \\
$G=8$ & \textbf{78.6} & \textbf{80.3} & \textbf{74.1} & \textbf{76.7} & \textbf{77.20} \\
\midrule
\multicolumn{6}{c}{\textit{Event Length} $L$} \\
\midrule
$L=1$ & 76.9 & 78.2 & 65.2 & \textbf{74.6} & 73.90 \\
$L=3$ & \textbf{77.5} & \textbf{78.2} & \textbf{71.6} & 73.4 & \textbf{74.61} \\
$L=5$ & 72.8 & 72.5 & 64.7 & 73.2 & 71.40 \\
\midrule
\multicolumn{6}{c}{\textit{Similarity Model}} \\
\midrule
VideoCLIP-XL \cite{videoclip} & \textbf{77.5} & \textbf{78.2} & \textbf{71.6} & 73.4 & \textbf{74.61} \\
ViCLIP \cite{viclip} & 76.3 & 75.6 & 66.2 & 73.6 & 73.01 \\
CLIP-large \cite{clip} & 77.5 & 76.2 & 68.2 & \textbf{74.4} & 74.24 \\
\midrule
\multicolumn{6}{c}{\textit{Similarity Reward}} \\
\midrule
w/ $r_s$ & 77.5 & 78.2 & 71.6 & 73.4 & 74.61 \\
w/o $r_s$ & 73.4 & 73.6 & 66.7 & 73.0 & 72.00 \\
\bottomrule
\end{tabular}
\end{table}

\textbf{Ablation of Event Chain Length.} 
We explore different values of $L$ to quantify the effect of event-chain length and granularity on prediction accuracy. The findings in \cref{tab:ablation} indicate a non-monotonic relationship: both extremes—too short and too long—are detrimental. Short chains fail to capture enough visual detail, while long chains introduce redundancy that complicates contextual reasoning. \\
\textbf{Ablation of Similarity Model.} We compare different schemes for computing the similarity reward $r_s$. Specifically, we evaluate the use of video-text similarity models, VideoCLIP-XL and ViCLIP, as well as the image-text similarity model CLIP. The comparable performance across these models demonstrates the robustness of our proposed approach, with VideoCLIP-XL yielding the best results. \\
\textbf{Ablation of Similarity Reward.} To test the impact of the similarity reward $r_s$ on model performance, we conducted training without the similarity reward signal. 
The experimental results show a noticeable decline in all metrics, demonstrating the effectiveness of the similarity reward.

\begin{figure}[t]
    \centering
    \includegraphics[width=1.0\linewidth]{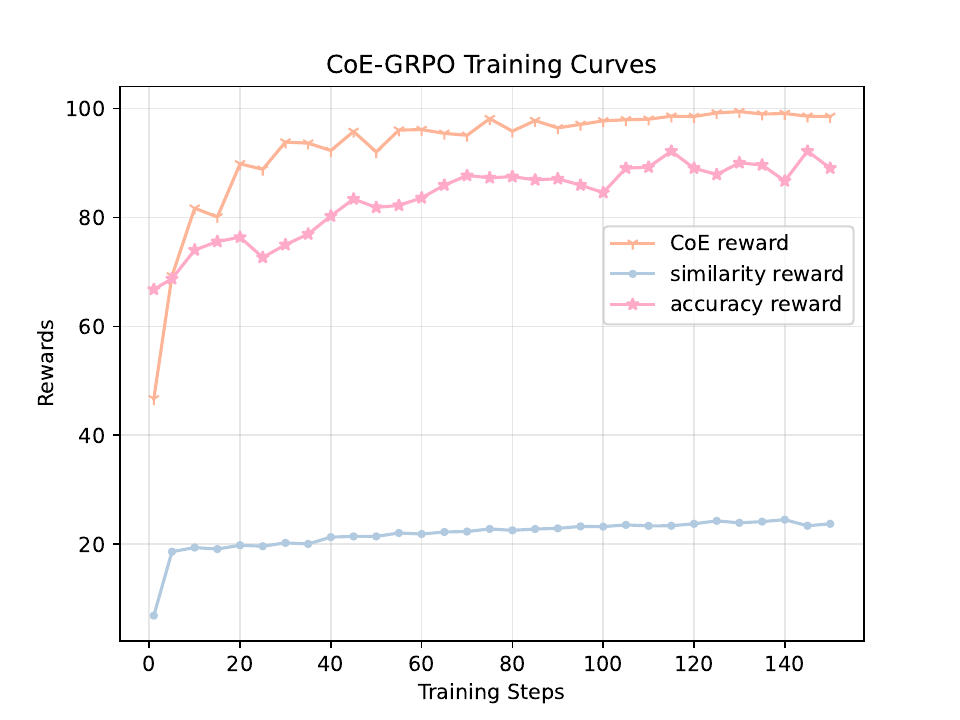}
    \caption{The training curves of CoE-GRPO.}
    \label{fig:traing_curves}
\end{figure}

\subsection{Training Curves}
As shown in  \cref{fig:traing_curves}, the accuracy reward $r_a$ exhibits a generally upward trend, indicating that the model continuously improves its event prediction ability under the training strategy. As for the curve of CoE reward $r_{e}$, the rapid increase observed during the initial stages suggests that the model learns the conceptual framework for constructing event chains. The subsequent steady ascent indicates that the model is gradually approximating the target event chain length defined by parameter $L$. The consistent increase in similarity reward $r_s$ indicates a progressive enhancement in the alignment between the event descriptions and the corresponding video segments. Notably, we found that even highly accurate descriptions yield similarity scores only in the range of $0.2-0.3$, indicating that the $r_s$ fall within a normal and reasonable range.

%% file: sec/5_conclusion.tex
\section{Conclusion}
Video event prediction holds significant practical value, yet research on MLLMs in this domain remains limited.
In this work, we present the first evaluation of various MLLMs on the video event prediction task, establishing comprehensive baselines. 
Through experiments, we reveal the reasons behind the inaccurate predictions in VEP, including the lack of logical reasoning ability for future events and insufficient utilization of visual information.
To address these challenges, we propose a \textbf{C}hain \textbf{o}f \textbf{E}vents (\textbf{CoE}) prediction paradigm, which unlocks the MLLMs' ability to construct temporal event chains and enables them to logically reason over the observed video to predict future events. Extensive experiments show that our proposed method can effectively overcome the challenges encountered by MLLMs in VEP, achieving state-of-the-art performance across existing benchmarks. We hope this work lays the foundation for future research on MLLMs to video event prediction.

%% file: sec/X_suppl.tex
\clearpage
\setcounter{page}{1}
\maketitlesupplementary
\setcounter{section}{0}
\renewcommand{\thesection}{\Alph{section}}
\section{Additional Results}
We additionally include comparisons with RL-based methods and traditional state-of-the-art approaches.
\begin{table}[h]
    \centering
    \small
    \caption{Additional results, *denotes traditional method.}
    \label{tab:additional_results}
    \renewcommand{\arraystretch}{0.3}
    \setlength{\tabcolsep}{3pt}
    \begin{tabular}{ccccc}
    \toprule
        \multirow{2}{*}{Model} & FutureBench $\uparrow$ & \multicolumn{3}{c}{AVEP $\uparrow$} \\
        \cmidrule{2-5}
         & AVG & Verb & Noun-F1 & Action-F1 \\
        \midrule
        VideoChat-R1 & 46.59 & 8.31 & 42.22 & 3.95 \\
        Video-R1 & 67.47 & 9.47 & 47.04 & 4.32 \\
        Ours & \textbf{75.00} & \textbf{18.75} & \textbf{64.03} & \textbf{9.88} \\
        \midrule
         EventFormer* & - & 22.71 & 46.24 & 7.69 \\
         \bottomrule
    \end{tabular}
    \vspace{-13pt}
\end{table} \\
\section{Attention Score Calculation Process}
To quantitatively evaluate the MLLMs' attention to visual information during the video event prediction task, we compute the attention weights assigned to both visual and text tokens. 
We record the attention score matrices of every head at every layer during inference on the test set. Since existing open-source MLLMs contain a large number of layers and heads, and we did not observe notable differences across them in our experiments, we visualize the average attention scores aggregated over all LLM layers and heads.
To control for the variation in the number of tokens generated by different models, we visualize the attention score distribution from option tokens to all other tokens. This provides a direct insight into how each model allocates attention between visual and textual information during event prediction, while also enabling an unbiased comparison of their visual attention. 

\section{Similarity Reward Computation}
For an event chain of length $n$, where the video segments are represented as $[clip_1,clip_2,\dots,clip_n]$ and the corresponding descriptions as $[\mathcal{D}_1,\mathcal{D}_2,\dots,\mathcal{D}_n]$, we use the similarity model $f_{\theta}(\cdot)$ to embed the video feature $v$ and text feature $t$ of the event chain:
\begin{equation}
    v_j=f_{\theta_{visual}}(clip_j),t_j=f_{\theta_{text}}(\mathcal{D}_j).
\end{equation}
Thus, the similarity reward $r_s$ is obtained by averaging the similarities between the video and text features of the events:
\begin{equation}
    r_s=\frac{1}{n}\sum_{j=1}^n(\text{sim}(v_j,t_j)),
\end{equation}
where $\text{sim}(\cdot)$ denotes the similarity computation function, which typically refers to cosine similarity.

In the experiments described in the main paper, we explore two different approaches for calculating the similarity reward. The first method directly uses a video-text alignment model to compute the video features of cropped video event segments and the text features of their corresponding descriptions. For this approach, we follow the official recommendation and use a frame rate of 8 frames for sampling.
The second method employs an image-text alignment model, where we extract image frames from the video segments, calculate the similarity between each image feature and the text feature, and then average these similarity scores to obtain the overall similarity between the video event segment and its textual description. We use a frame rate of 8 frames for sampling as well.

\section{Video Event Prediction Benchmarks}
\textbf{Futurebench.} 
Futurebench is a benchmark specifically designed to evaluate the video event prediction capability of MLLMs, featuring both SFT and GRPO training datasets. This benchmark collects video data from various perspectives, different lengths, and types, offering a comprehensive assessment of event prediction performance across diverse scenarios. To evaluate the event prediction ability of MLLMs from multiple angles, Futurebench categorizes the event prediction tasks into four types:
\begin{itemize}
    \item \textbf{1-Hop}: The model predicts a single future event that directly links the observed scenes to the final one, corresponding to a standard Next Event Prediction (NEP).
    \item \textbf{2-Hop}: The model infers a sequence of two consecutive future events, requiring a short chain reasoning process that sequentially connects the observed scenes to the final event.
    \item \textbf{3-Hop}: The model predicts three consecutive future events, significantly increasing task complexity by necessitating deeper causal reasoning across a longer temporal span.
    \item \textbf{Internp.}: The model must infer multiple non-consecutive future events, given a set of partially observed scenes that include intermediate anchor events.
\end{itemize}
In this benchmark, we construct \textbf{2,000} CoE-SFT samples for CoE-SFT training and additionally utilize the provided \textbf{2,000} reinforcement learning samples as training data for CoE-GRPO. As the training set contains only 1-Hop and 2-Hop event prediction samples, performance gains on the 3-Hop and Interp. metrics provide a reliable measure of the model’s generalization capability on VEP task.

\textbf{AVEP.} 
Action-centric Video Event Prediction (AVEP) is a benchmark specifically designed for evaluating the video event prediction capabilities of models. This benchmark primarily focuses on the events themselves, providing a comprehensive and fine-grained assessment of the model's performance on video event prediction. Video events are decomposed into event arguments, and the model's ability to predict future events is evaluated at the argument level. The evaluation of a model's event prediction performance on AVEP is based on the following key aspects:
\begin{itemize}
    \item \textbf{Verb Accuracy}: This metric measures the accuracy of the model in predicting the trigger verb of future events. As verbs are the core triggers of events, they are crucial components in event construction, and their prediction reflects the model's logical reasoning ability.
    \item \textbf{Noun Metrics}: This set of metrics assesses the model's ability to predict the participants in future events, including the subject, object, and tool. These metrics reflect the model's consistency in role prediction, evaluating the logical coherence of the event arguments.
    \item \textbf{Action Metrics}: These metrics evaluate the noun prediction performance when the verb is correctly predicted, providing a direct indication of the model's video event prediction ability.
\end{itemize}
The AVEP dataset provides the ground truth but does not include the SFT data required for training MLLMs. Therefore, we use the \textbf{5,000} constructed CoE-SFT samples for supervised fine-tuning and experiments.
In this benchmark, we select \textbf{5,000} samples from the provided dataset as the training data for GRPO and CoE-GRPO, and evaluate the model on the entire validation and test sets of the benchmark.

\begin{figure}[t]
    \centering
    \includegraphics[width=1\linewidth]{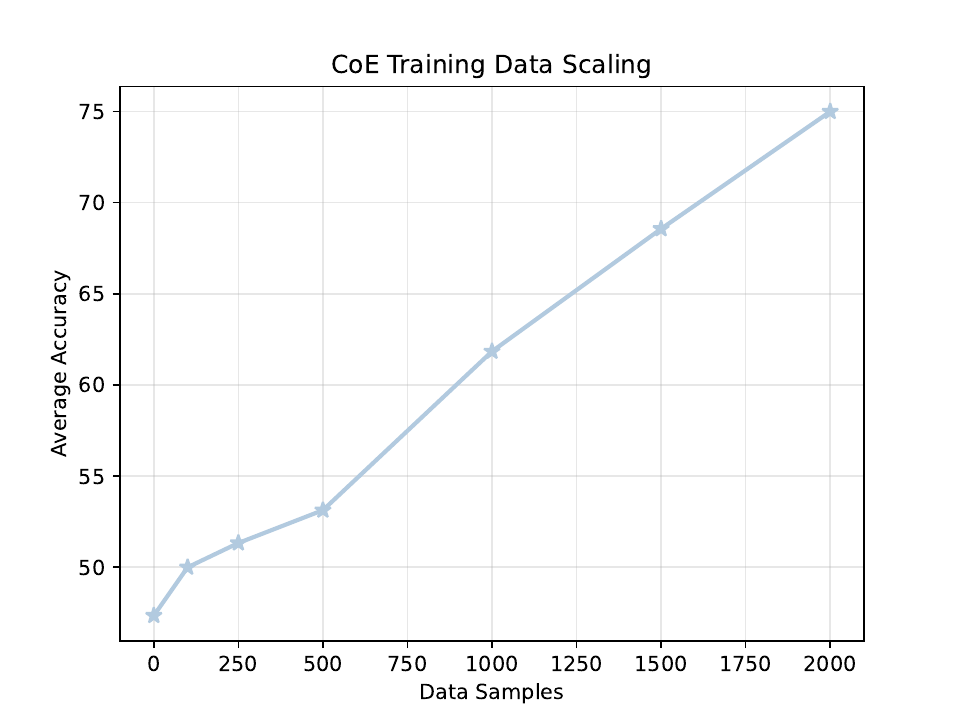}
    \caption{The data scaling curve of CoE}
    \label{fig:data_scaling}
\end{figure}

\section{Training Data Scaling of CoE}
Using the benchmark-provided training data, we train the model to follow the CoE paradigm, and observe strong improvements on VEP task, highlighting the data efficiency of our method.

To further investigate the effect of data scale, we conduct a data-scaling study. As shown in the \cref{fig:data_scaling}, directly applying the CoE paradigm without any CoE-specific training leads to a performance drop, indicating that the model is unable to effectively leverage visual information to construct the logical connections to future events.
When a small amount of training data is provided, performance begins to improve steadily, suggesting that the model is gradually acquiring the CoE reasoning pattern. As the data size increases, performance improves rapidly, demonstrating the strong data efficiency of our approach.
Notably, although we train the model on only 2,000 samples from FutureBench, the upward trend in the scaling curve remains far from saturated. This indicates that additional data would likely yield further gains, highlighting the strong scalability and continued potential of our proposed method .

\section{Details of Judge Model Evaluation}
To more accurately assess MLLMs' event prediction capabilities in real-world applications, the judge model evaluation is designed to reflect an open-set prediction setting. Specifically, we remove all answer options and require the model to directly reason about the observed video and predict the most plausible future event. In the evaluation, we provide the judge model with the video, question, and reference answer. The judge model then evaluates each output from two perspectives: (i) the logical consistency and soundness of the reasoning, and (ii) the correctness of the predicted event.
Directly scoring model outputs may lead to inconsistencies or hallucination from the judge model, undermining evaluation fairness. To address this, we employ a group-wise comparison protocol and report the win rate as the evaluation metric, which yields a more reliable assessment.

\begin{figure*}
    \centering
    \includegraphics[width=1\linewidth]{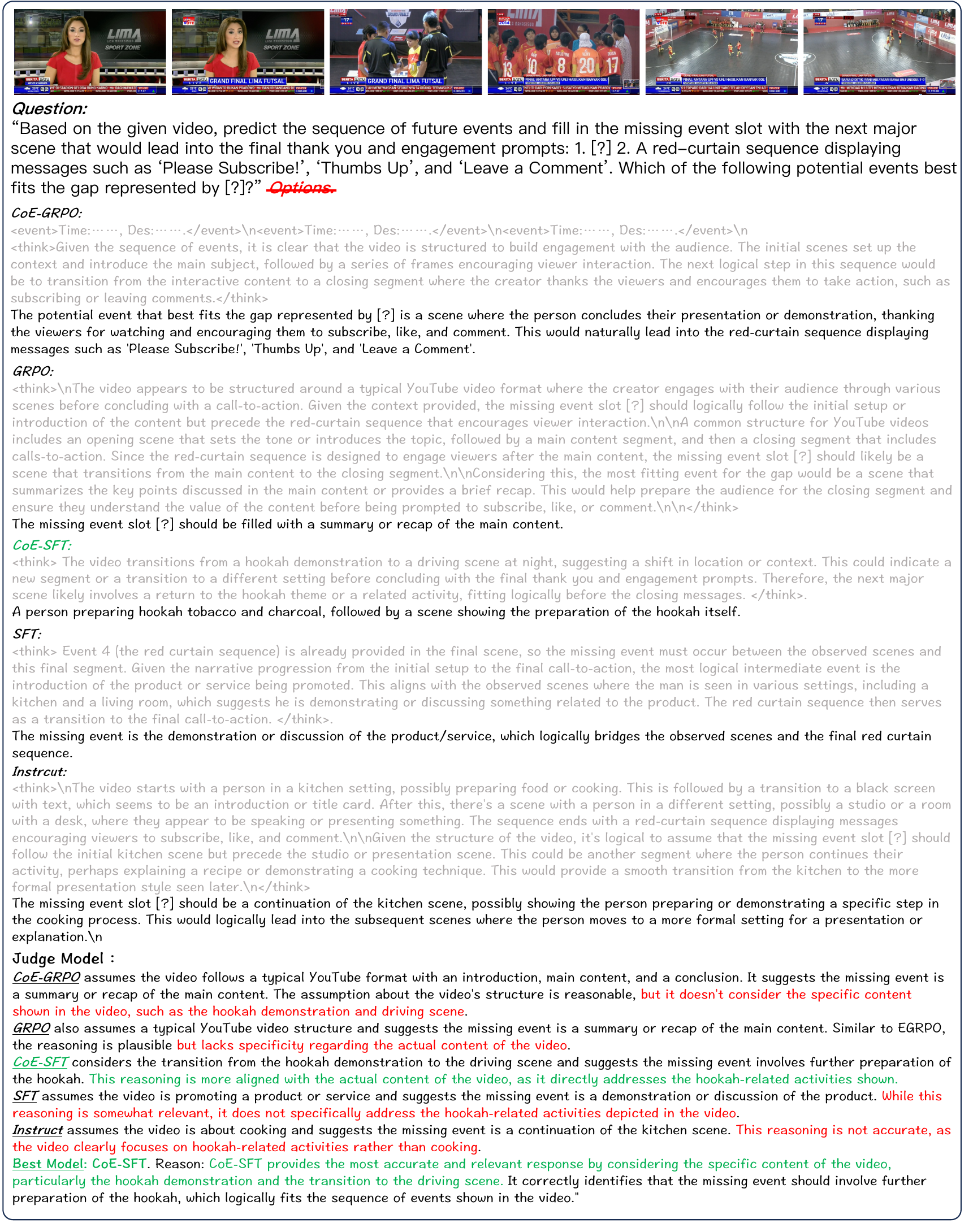}
    \caption{An example of judge model evaluation.}
    \label{fig:judgment_full}
\end{figure*}

We provide a comprehensive example of the judge model evaluation, as shown in \cref{fig:judgment_full}. 
In this evaluation, it is observed that the models trained with the CoE-SFT method produce reasoning processes that are more closely aligned with the video content, and their reasoning is visually grounded, clear, and concise. In contrast, other methods either fail to focus on the visual content during reasoning or provide predictions that lack logical consistency. In other test samples, we find that both CoE-SFT and CoE-GRPO methods are able to deliver reasonable reasoning processes and accurate predictions in the absence of options.

\section{Examples of CoE}
\begin{figure*}
    \centering
    \includegraphics[width=1\linewidth]{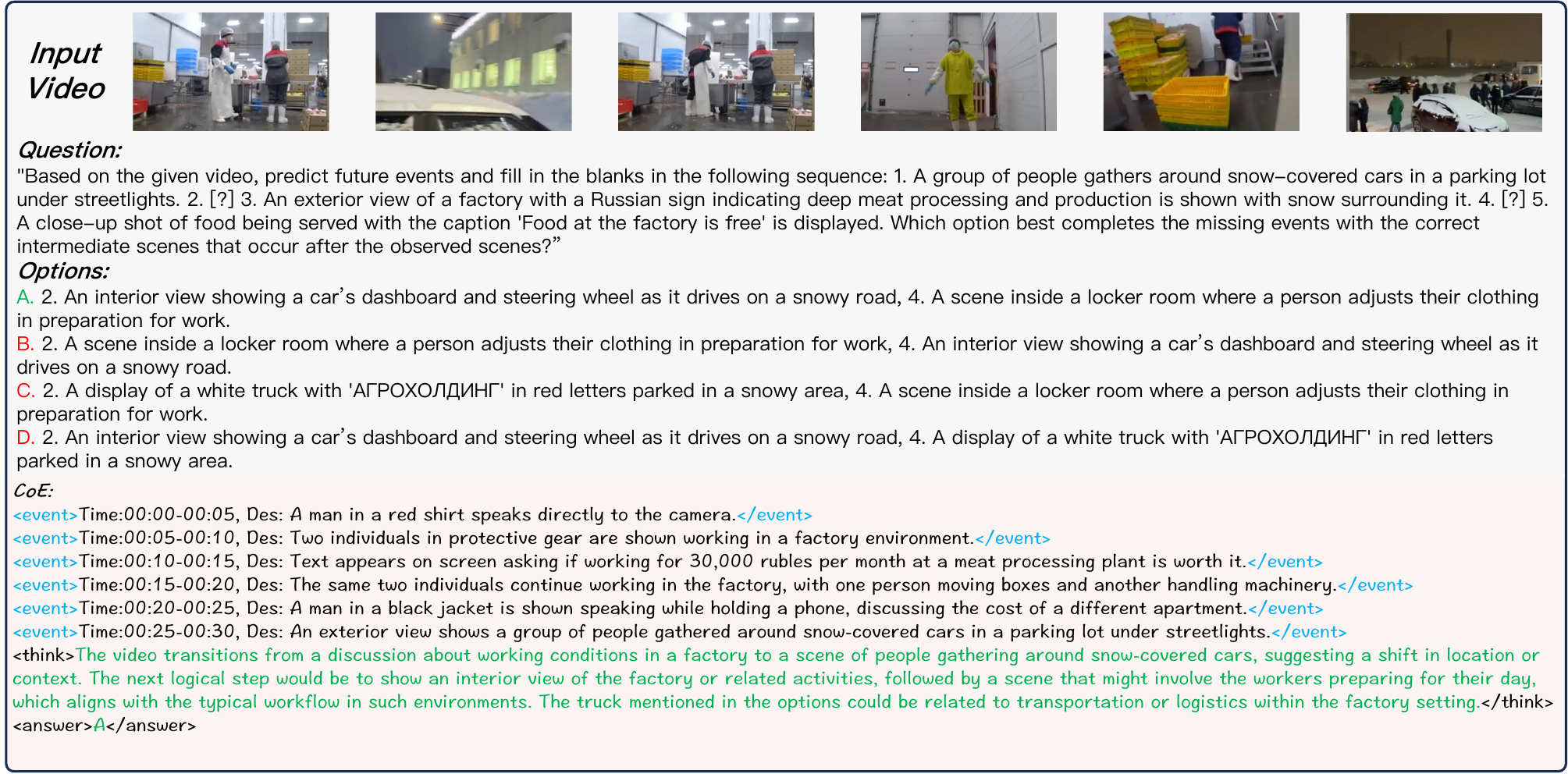}
    \caption{An example of CoE paradigm.}
    \label{fig:example_0}
\end{figure*}

\begin{figure*}
    \centering
    \includegraphics[width=1\linewidth]{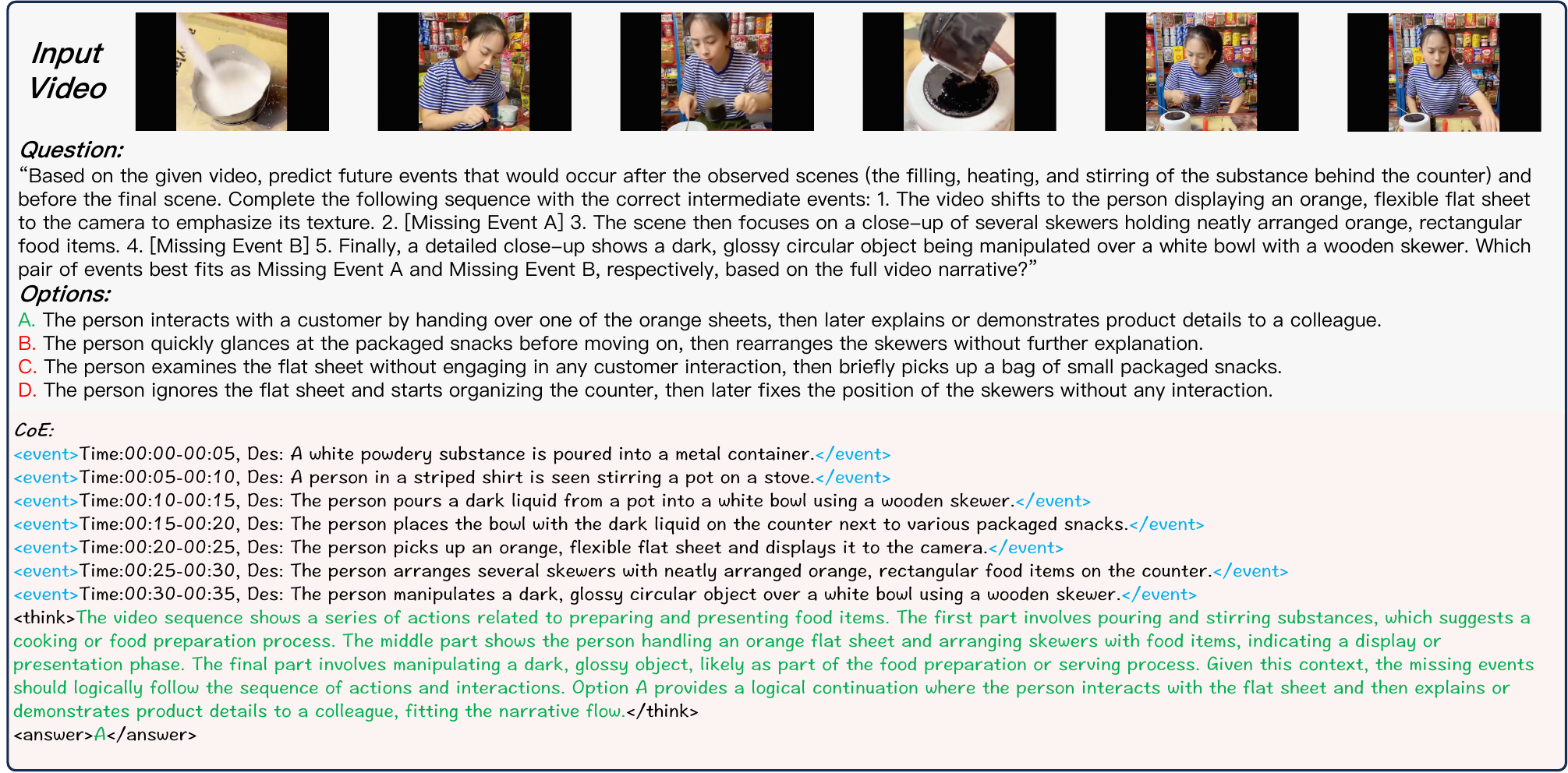}
    \caption{An example of CoE paradigm.}
    \label{fig:example_1}
\end{figure*}

\begin{figure*}
    \centering
    \includegraphics[width=1\linewidth]{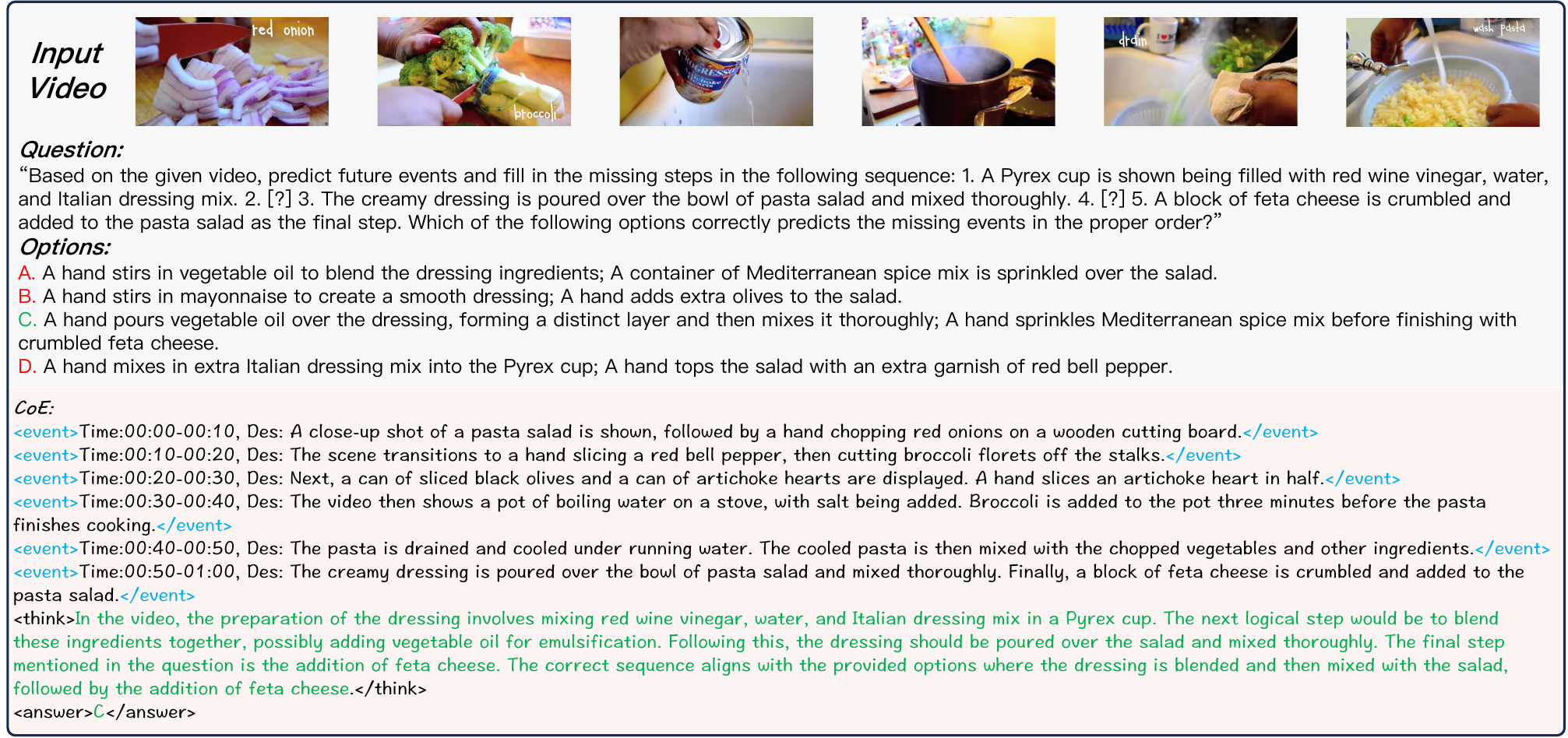}
    \caption{An example of CoE paradigm.}
    \label{fig:example_2}
\end{figure*}

\begin{figure*}
    \centering
    \includegraphics[width=1\linewidth]{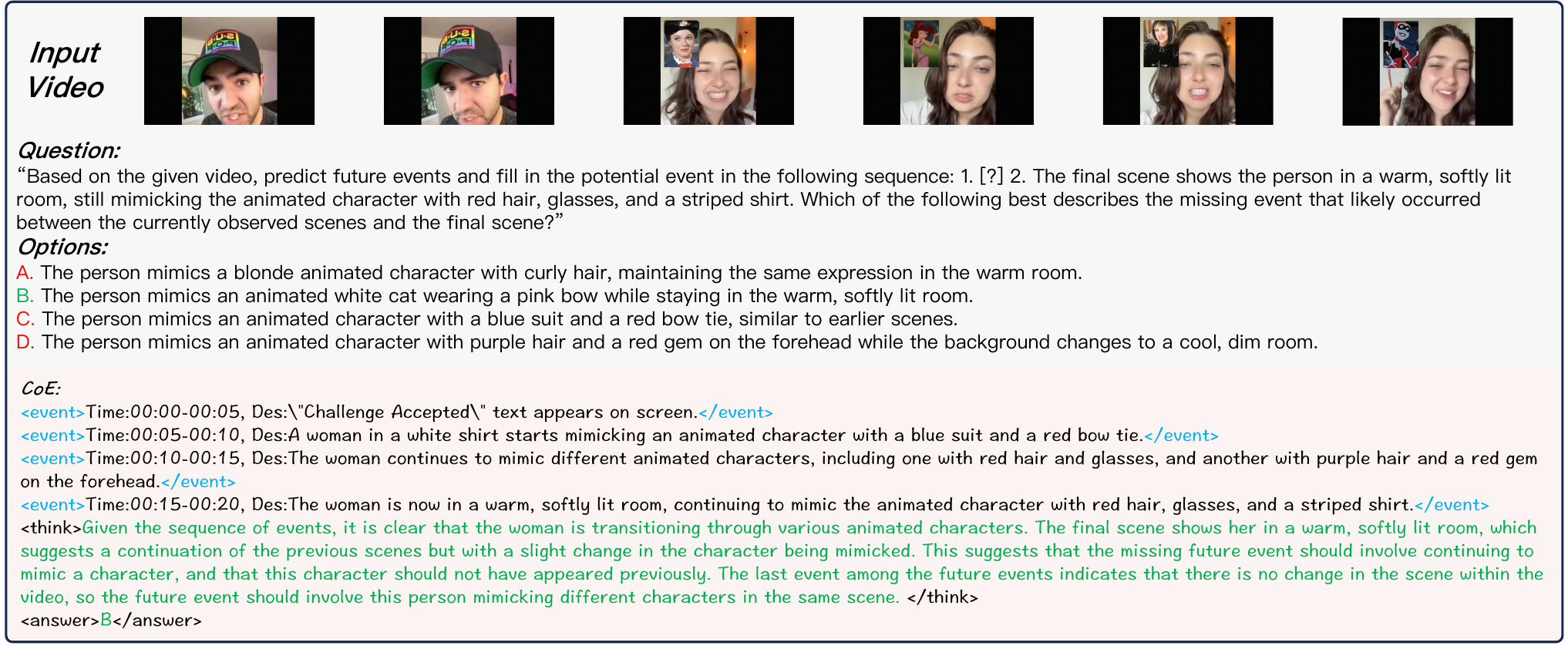}
    \caption{An example of CoE paradigm.}
    \label{fig:example_5}
\end{figure*}

\begin{figure*}
    \centering
    \includegraphics[width=1\linewidth]{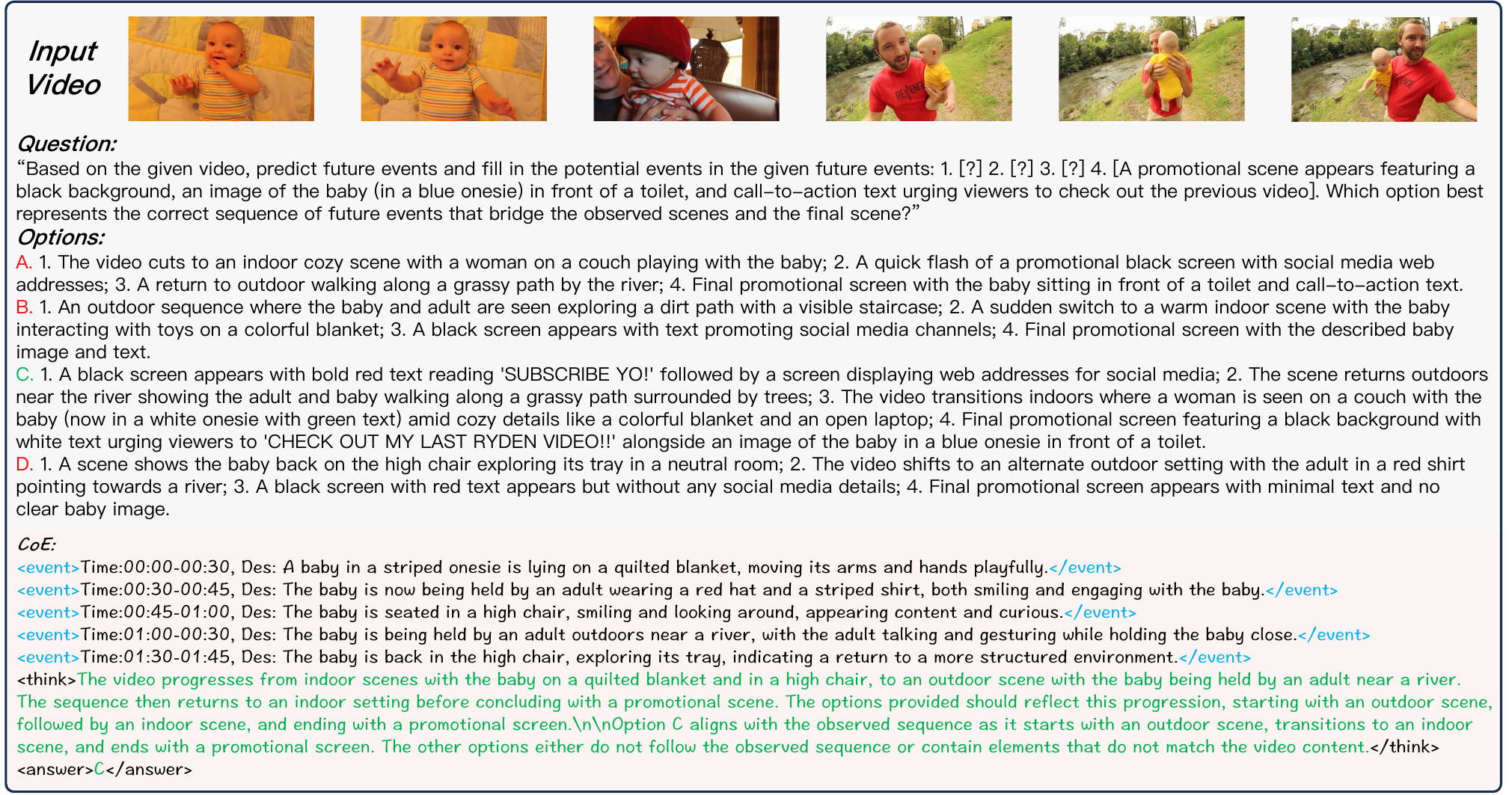}
    \caption{An example of CoE paradigm.}
    \label{fig:example_3}
\end{figure*}

\begin{figure*}
    \centering
    \includegraphics[width=1\linewidth]{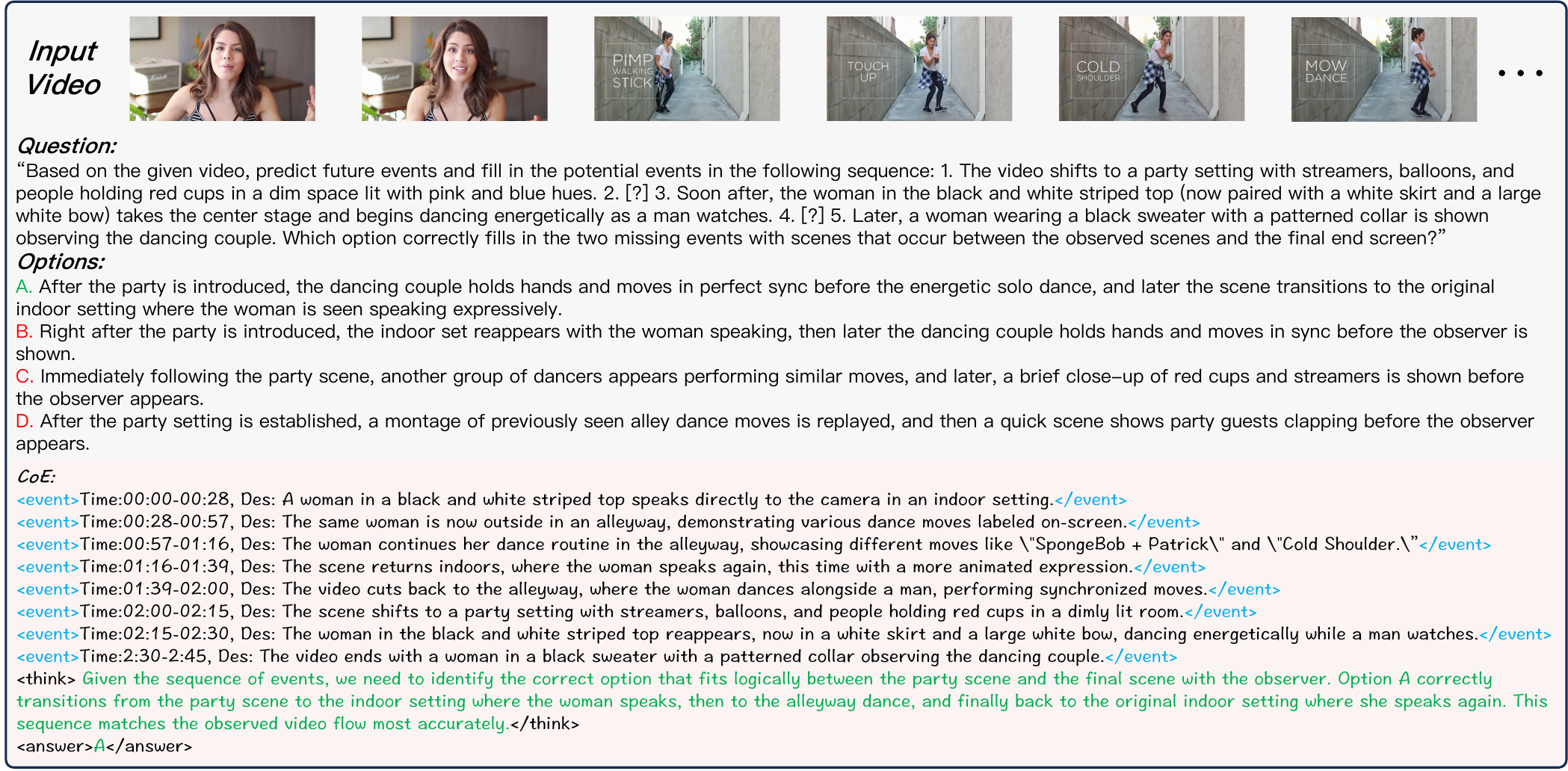}
    \caption{An example of CoE paradigm.}
    \label{fig:example_4}
\end{figure*}

As illustrated in \cref{fig:example_0,fig:example_1,fig:example_2,fig:example_3,fig:example_4,fig:example_5}, we present several randomly selected prediction examples generated by our proposed method.
From the constructed event chains in these results (marked as blue in the figures), it can be observed that our approach enables the model to perform fine-grained temporal modeling of the input videos and efficiently improves the utilization of viusal information.
Notably, as the overall video duration varies, the granularity of event segmentation adaptively adjusts, while the length of the generated event chains remains relatively stable.
In addition, the textual descriptions produced for each event are generally consistent with the   corresponding video segments.

The followed reasoning process (marked as green in the figures) exhibited in these examples demonstrates that the model can logically infer future events based on the details present in the video. This not only shows that our proposed method enables the model to effectively establish logical connections between the video and future events, but also indicates that after CoE-GRPO training, the model retains the logical reasoning capabilities learned during CoE-SFT.

\begin{figure*}
    \centering
    \includegraphics[width=1\linewidth]{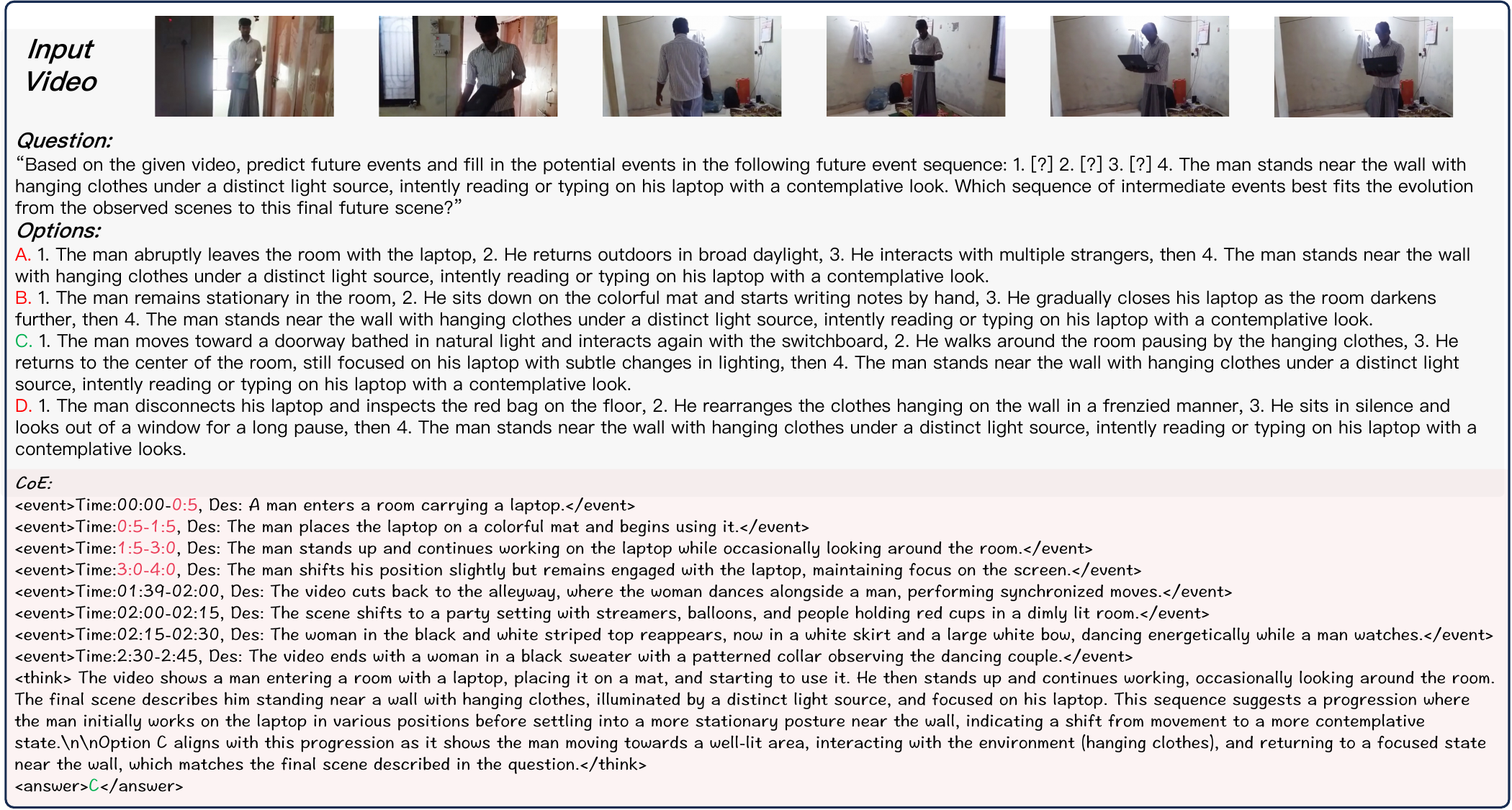}
    \caption{A bad case of CoE paradigm.}
    \label{fig:bad_case_0}
\end{figure*}

\begin{figure*}
    \centering
    \includegraphics[width=1\linewidth]{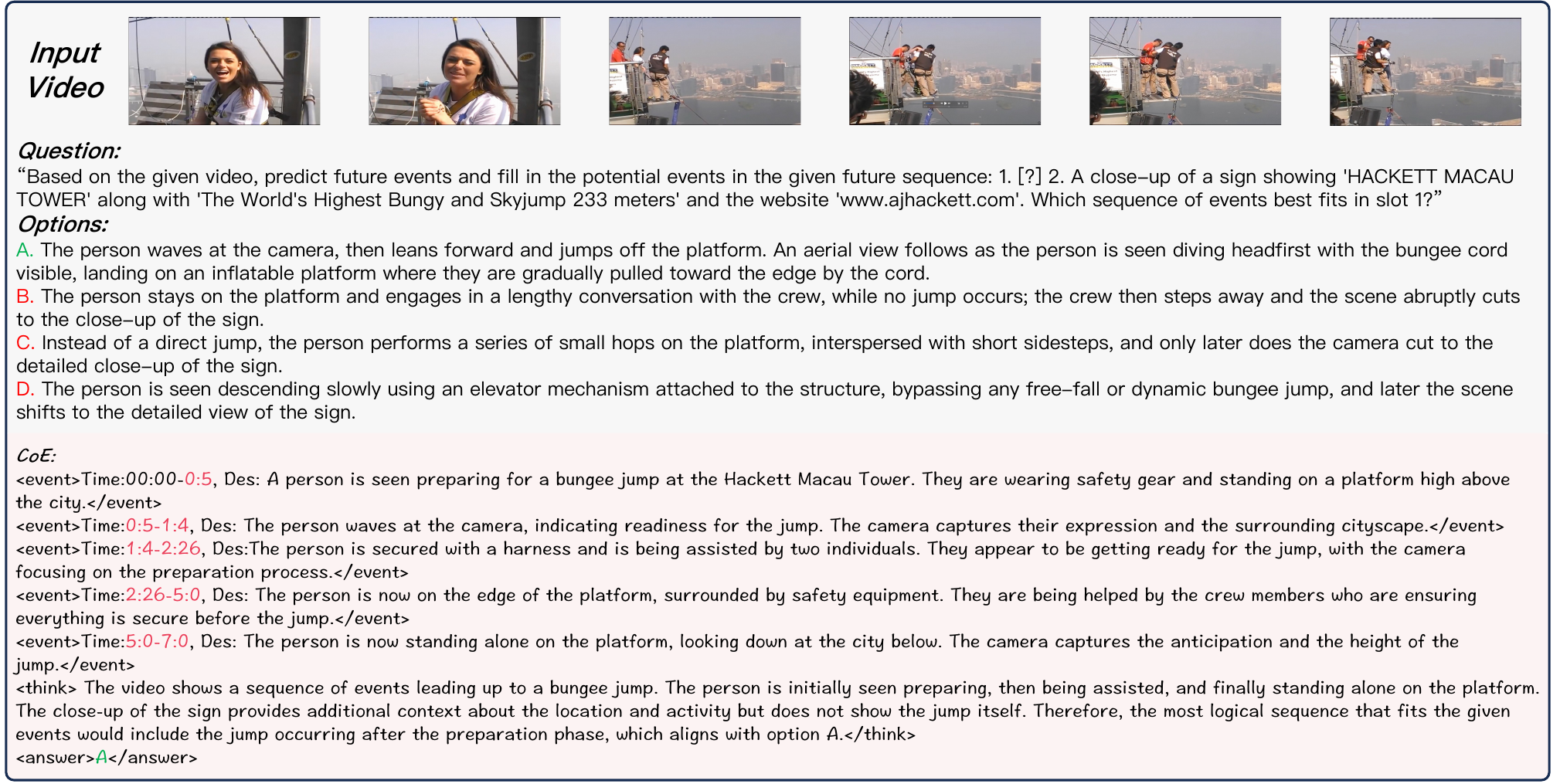}
    \caption{A bad case of CoE paradigm.}
    \label{fig:bad_case_1}
\end{figure*}

However, as illustrated in the \cref{fig:bad_case_0,fig:bad_case_1}, we also observed some bad cases, though their occurrence is extremely rare—approximately three instances out of one thousand samples. In these cases, the model fails to generate timestamps correctly according to the given instructions. Nevertheless, this issue has minimal impact on the model’s event descriptions and prediction results.

\section{Prompt Templates}
\begin{figure*}
    \centering
    \includegraphics[width=1\linewidth]{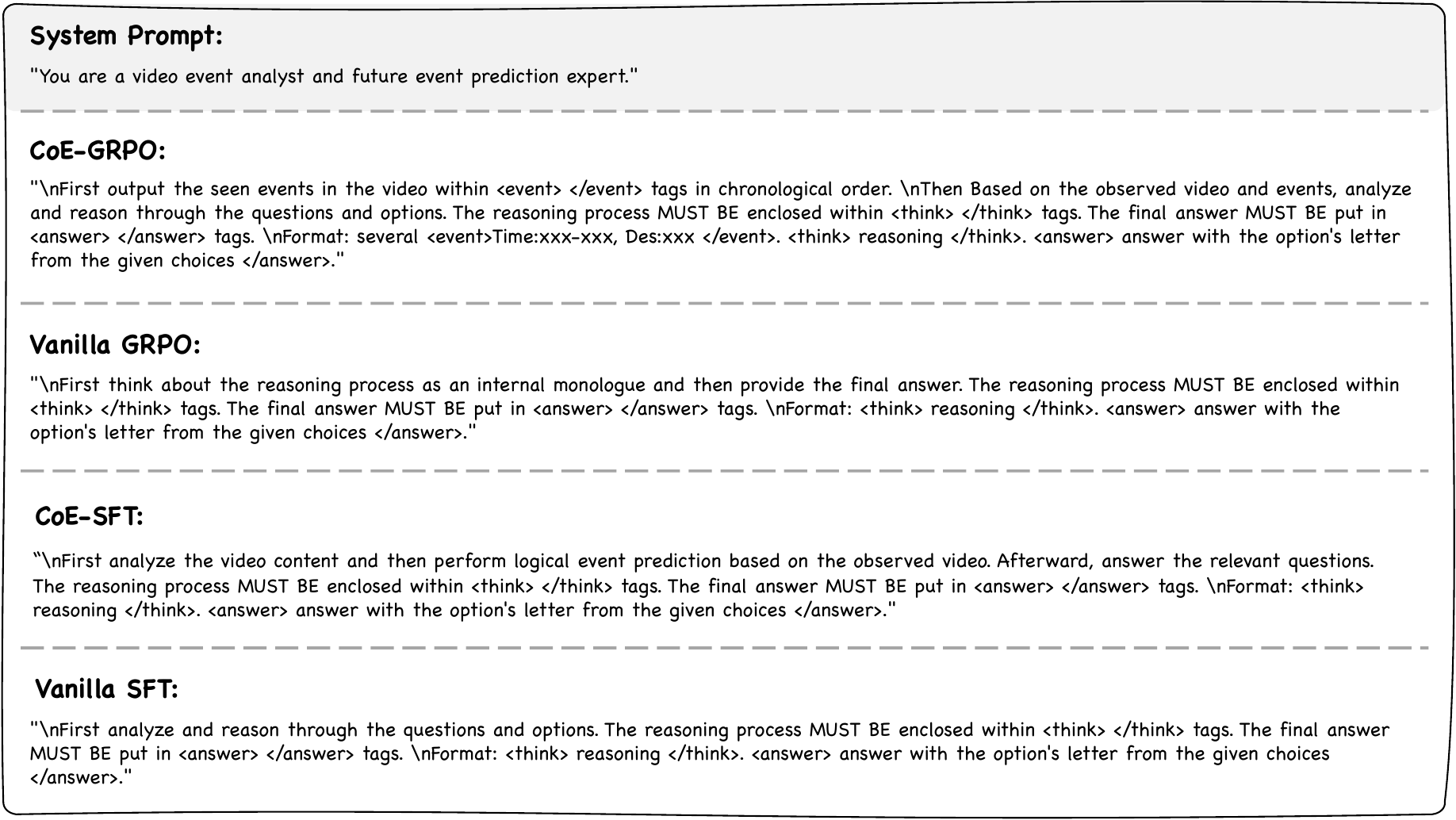}
    \caption{Prompt template for training and inference.}
    \label{fig:prompt_0}
\end{figure*}

\begin{figure*}
    \centering
    \includegraphics[width=1\linewidth]{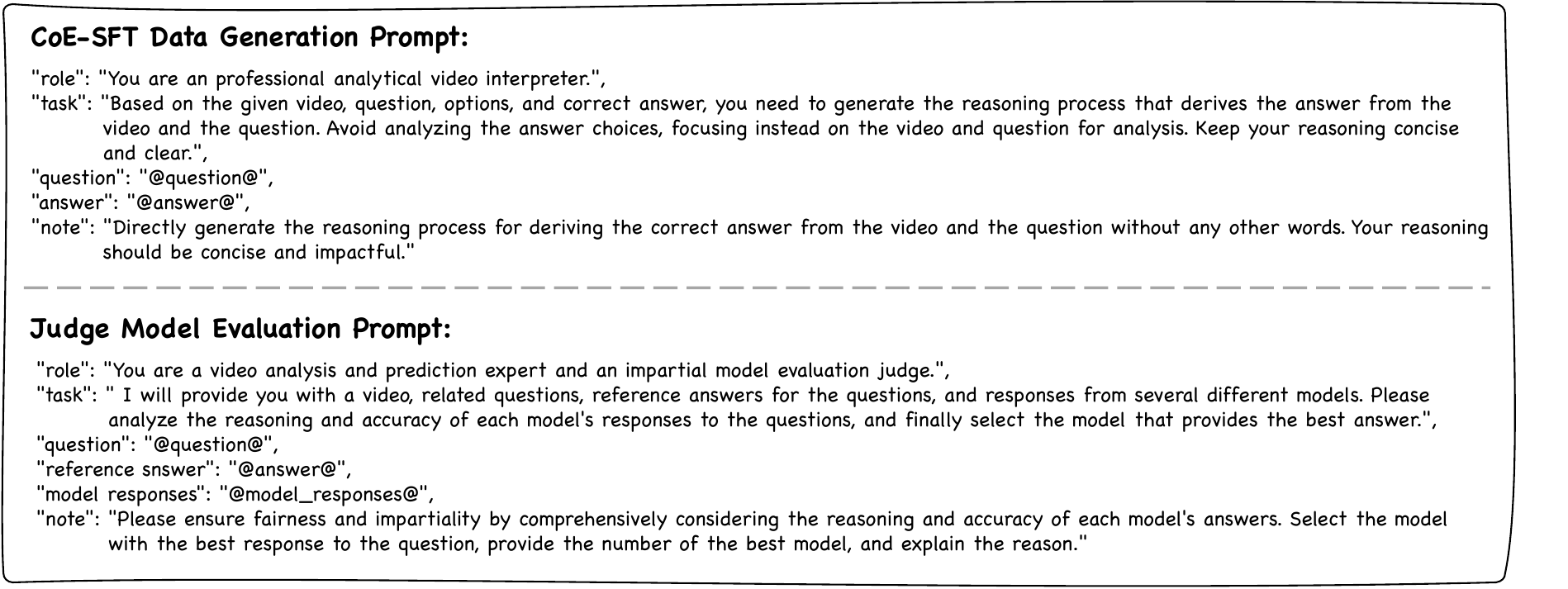}
    \caption{Prompt template for CoE-SFT data generation and judge model evaluation.}
    \label{fig:prompt_1}
\end{figure*}

\cref{fig:prompt_0} illustrates the prompt template we use for training and inference of all MLLMs. We also present the prompt used for CoE-SFT data generation and judge model evaluation in \cref{fig:prompt_1}

\section{Visualizations of Attention Increase in Visual Tokens}
\begin{figure*}
    \centering
    \includegraphics[width=1\linewidth]{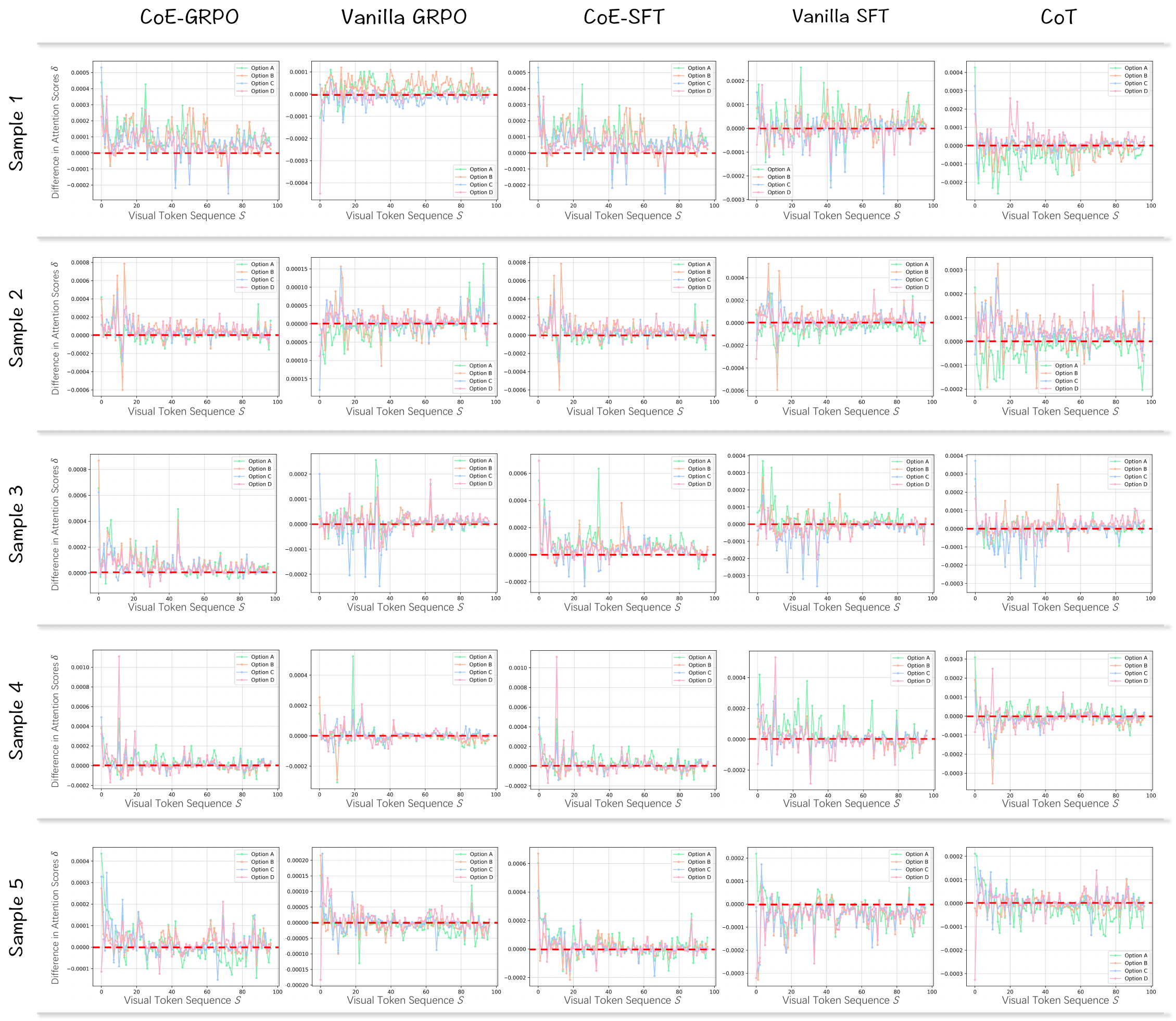}
    \caption{Some examples of attention differences comparing to the vanilla model.}
    \label{fig:attn_diffs}
\end{figure*}

Here are some visualizations of the attention differences between the post-trained models and the vanilla model (Qwen2.5-VL-7B-Instruct) for visual tokens, as shown in \cref{fig:attn_diffs}. The portions of the curves above 0 indicate an increase in attention. It is clear that both the CoE-GRPO and CoE-SFT methods effectively enhance the model’s focus on visual information. However, while both vanilla GRPO and vanilla SFT improve the model’s event prediction performance, they fail to adequately address the issue of insufficient visual information utilization, thus limiting the overall efficiency of performance improvement. Additionally, when we directly prompt the model to predict future events in a CoT manner without training, it does not significantly improve the model's attention to visual content.

\section{Limitations and Future Works}
We hope that this work can serve as a foundation and provide inspiration for future explorations of MLLMs’ capabilities in video event prediction. Below, we also summarize the limitations and outline several potential directions for future research:
\begin{itemize}
    \item \textbf{Temporal Localization Ability.} Our proposed method imposes certain requirements on the model’s temporal localization capability. Although most current MLLMs demonstrate strong temporal localization performance, there is still considerable room for improvement in accuracy. We believe that further enhancement of this capability could lead to additional performance gains for our method.
    \item \textbf{Structures of Historical Events.} We explored the use of event-chain construction to strengthen the model’s fine-grained temporal modeling of videos. However, we only investigated relatively basic forms of event chains. More complex formulations for modeling historical event structures—such as relation-aware event chains or event graphs—remain promising directions for future research. 
    \item \textbf{Perfomance on Other MLLMs and Tasks.} Since our approach does not impose specific requirements on the base model, we believe that the CoE paradigm should be applicable to all MLLMs. However, due to resource constraints, we were unable to conduct experiments across all available models. Nevertheless, the exceptional performance of our method on VEP demonstrates its potential for application to other tasks as well. Exploring the applicability of CoE to other video-related tasks presents an interesting avenue for future research.
\end{itemize}

%